%% file: HyperODE_arxiv.tex
\documentclass[letterpaper]{article} 
\usepackage{aaai2027}  
\nocopyright  
\usepackage[hyphens]{url}  
\usepackage{graphicx} 
\usepackage{natbib}  
\usepackage{caption} 
\usepackage{algorithm}
\usepackage{algorithmic}
\usepackage{amssymb}
\usepackage{newfloat}
\usepackage{listings}
\DeclareCaptionStyle{ruled}{labelfont=normalfont,labelsep=colon,strut=off} 
\floatstyle{ruled}
\newfloat{listing}{tb}{lst}{}
\floatname{listing}{Listing}

\usepackage{booktabs}
\usepackage{multirow}
\usepackage{amsmath}
\usepackage{amsthm}
\usepackage{subcaption}

\title{HyperODE: Zero-Shot Surrogate for Simulation and\\ Inference of Dynamical Systems}
\author{
    Ajitesh Srivastava
}
\affiliations{
    Northeastern University\\
    aj.srivastava@northeastern.edu
}

\newcommand{\gnnFac}{5}            
\newcommand{\hoFourFac}{1.4}       
\newcommand{\forcingCov}{0.91}     

\begin{document}

\maketitle

\begin{abstract}
Understanding and controlling complex dynamical systems often requires executing thousands of numerical simulations across vast parametric landscapes, which is time-consuming. Machine learning surrogates significantly accelerate simulation by predicting state trajectories across different initializations and parameter values. However, surrogate models are specialized to one simulation model. Modifying the underlying differential equations - e.g., adding a physiological state or altering an epidemiological contact network - renders trained models obsolete and forces computationally expensive retraining from scratch. We introduce HyperODE, a surrogate capable of operating across an entire class of approximately mass-conserving compartmental models without retraining. By mapping the structure of ordinary differential equations (ODEs) into directed hypergraphs, HyperODE decouples the functional form of system interactions from the neural network architecture. HyperODE takes a compartmental model in the form of an ODE with an arbitrary parameter distribution defined through quantiles and transforms it into a hypergraph. It builds a hypergraph neural ODE, which results in the distribution of the trajectories for all the states in the original ODE in the form of quantiles. We then use this surrogate to build an encoder that takes a noisy trajectory and outputs a distribution over the parameters of the original ODE, thus calibrating the model in a single forward pass instead of traditional expensive Markov Chain Monte Carlo (MCMC) sampling. On model families and system sizes never seen in training, HyperODE produces calibrated quantile bands in a single forward pass, with weighted-interval score and coverage on par with specialized surrogates for each structure. For inverse inference, HyperODE produces calibration from noisy state trajectories in a few milliseconds with a single shared encoder, competitive with existing methods. HyperODE extends zero-shot to ODEs that break mass conservation and to external forcing.
\end{abstract}


\section{Introduction}

Numerical simulation of parameterized dynamical systems is fundamental to much of computational sciences. Scenario modeling in an epidemic, screening a chemical reaction network, or designing a control policy all reduce to solving a system of ordinary differential equations (ODEs) thousands of times, sweeping over uncertain parameters and initial conditions. This process takes a considerable amount of time and computational resources, and accelerating it can help advance scientific discovery.
Machine learning surrogates aim to achieve this by replacing the numerical solver with a fast neural forward pass mapping parameters and initial conditions to state trajectories~\cite{lu2021deeponet,li2020fourier,sanchezgonzalez2020learning,pfaff2021learning,kochkov2021machine}. Existing surrogates, however, are almost always tied to a \emph{single, fixed} model. Even a small change in the governing equations -- e.g., adding a latent compartment to an epidemic model, rewiring a contact or mobility network, or introducing a new reaction channel, the learned surrogate is invalidated and must be retrained from scratch, at the cost of regenerating training data and re-optimizing the network. This prevents surrogates from being reused across the many closely related models that arise in practice.

\begin{figure}[!h]
\centering
\includegraphics[width=\columnwidth]{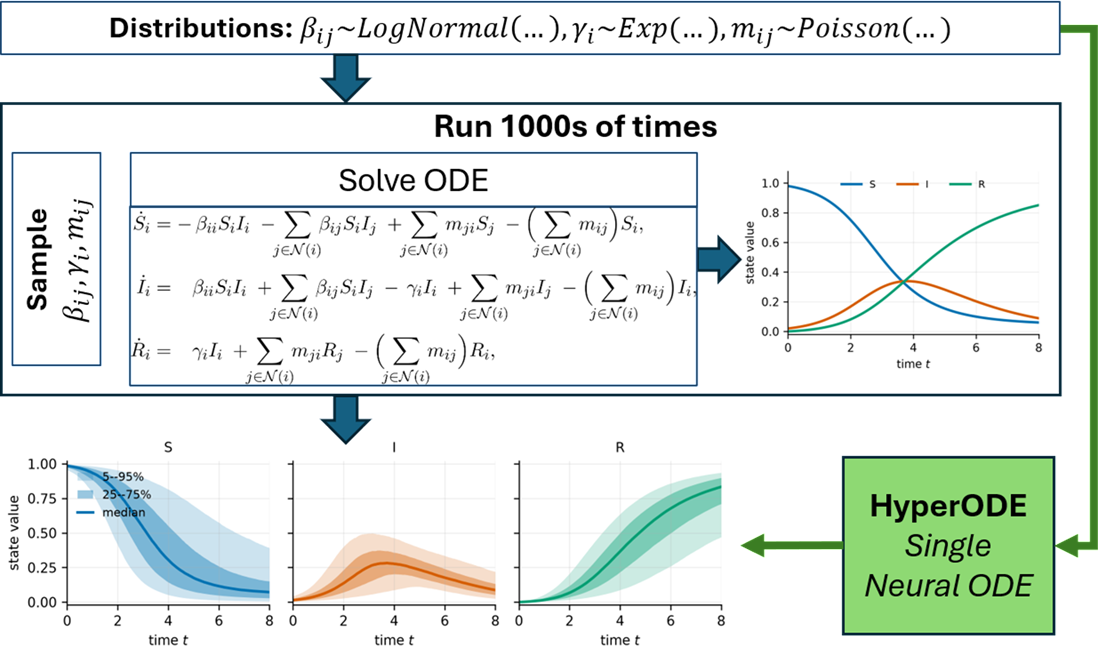}
\caption{Simulating a meta-population SIR model over $k$ coupled groups.}
\label{fig:problem}
\end{figure}
We instead seek a \emph{single} surrogate that works across an entire \emph{class} of models with no retraining: given an ODE with distribution over parameters it should return calibrated uncertainty on the resulting trajectories (not a point estimate). For example, consider the meta-population SIR epidemic model over $k$ coupled groups, with susceptible $S_i$, infected $I_i$, and recovered $R_i$ states in each group $i=1,\dots,k$ (Figure~\ref{fig:problem}).
For simulating this with an initial condition, one may assume some distributions for the parameters $\beta_{ij}$, $\gamma_i$, and $m_{ij}$, and then sample from these distributions to generate a set of trajectories -- requiring one ODE solve per sampled parameter set. The result is a distribution over trajectories quantified by trajectory quantiles.
Alternatively, given a single observed trajectory, one may want to calibrate the models by inferring the parameters, typically done with MCMC sampling, which requires many ODE solves to explore the posterior distribution over parameters~\cite{tan2022statistical}. Both processes are expensive, especially for large $k$.

Our key observation is that the structure of an ODE can be exposed to a neural network as a \emph{directed hypergraph}, in which each node is a state variable and each hyperedge is one additive monomial term of the dynamics, describing the interactions between the states. A single hyperedge operator, shared across every node and edge of every system defines the vector field. 
We propose \emph{HyperODE} -- a single pretrained model (with less than $15$k parameters) predicts quantile trajectory bands for an unseen system in one forward pass. Being a fully differentiable model, it enables gradient-based calibration to recover a parameter distribution from a single noisy trajectory. HyperODE can be further developed into a single-pass calibration approach that uses an encoder to consume a noisy trajectory, produces parameter distributions, and reuses the surrogate to construct quantile trajectory bands in milliseconds. The same surrogate and calibration model generalize zero-shot across a wide range of ODE families, system size, and interaction order. HyperODE is designed for inference on approximately mass conserving systems such as meta-population compartmental models and reaction networks. 
Specifically, our contributions are:
\begin{enumerate}
  \item We propose a hypergraph operator, agnostic to system structure, scale, and interaction order, that generates the trajectory distribution of ODEs, given the structure and a distribution over parameters.
  \item We show that the surrogate can be utilized for gradient-based calibration from noisy input trajectory.
  \item We propose an encoder that  when attached to the surrogate can directly generate a calibration distribution from a single noisy trajectory in one forward pass. It trades off accuracy for speed.
  \item We demonstrate that it applies to external forcing (seasonality, time-varying mobility) that reuses the pretrained model without retraining.
\end{enumerate}

\section{Background and Related Work}

\paragraph{Simulation and Machine Learning surrogates.}
Across the sciences, understanding and controlling a system means simulating its governing equations, often ordinary differential equations, repeatedly across many parameter settings and initial conditions. As an example, for performing scenario projection for an ongoing epidemic, a common approach is as follows~\cite{borchering2021modeling}. 
We sample model parameters from distributions and solve an ODE thousands of time to get multiple trajectories and then generate quantile bands to present distribution of the outcome. These thousands of ODE solves constitute the bottleneck for calibration, uncertainty quantification, and control. Machine-learning surrogates attempt to reduce this cost -- a trained neural network maps parameters and initial conditions directly to trajectories in a single fast forward pass, replacing the numerical solver.
Neural operators such as the Fourier neural operator and DeepONet learn the solution map of a differential equation directly, so that a single forward pass returns the solution for a new parameter field in place of a numerical solve \citep{li2020fourier,lu2021deeponet}. Graph-network simulators instead learn the local update rule of a physical system from data, rolling out particle or mesh dynamics faster than the simulator they imitate \citep{sanchezgonzalez2020learning,pfaff2021learning}. A complementary line embeds learned components inside the numerical method itself, accelerating a classical solver while preserving its stability and accuracy guarantees \citep{kochkov2021machine}. However, every such surrogate is trained for one fixed system, so any change to the governing equations -- adding a compartment, rewiring a contact network, introducing a reaction -- makes the trained surrogate obsolete and forces retraining. We aim to design a \emph{single} surrogate that can serve an entire \emph{class} of systems, transferring across  structures and scales without retraining.


\paragraph{Inverse Problems and Calibration.}

Another critical task is to calibrate a given ODE from noisy observed data. As an example, for performing scenario projections of an ongoing epidemic, we first fit the model parameters to noisy observations accounting for uncertainty~\cite{borchering2021modeling}, so that we can alter them to emulate a future scenario and run the forward simulations. Complex models are often unidentifiable -- distinct set of parameters can produce identical results. Therefore, the quality of calibration is measured by how well the quantile bands generated by simulating with the inferred parameters cover the observed noisy trajectories. 
Recovering parameters from observations is classically posed as Bayesian calibration and solved by MCMC or variational inference on the true simulator -- accurate but expensive and, as we show, prone to poor mixing at scale. Amortized simulation-based inference instead trains a neural estimator to map an observation to a posterior in one forward pass, such as neural posterior and neural ratio estimation \citep{papamakarios2016fast,greenberg2019automatic,hermans2020likelihood}. Crucially, these estimators are trained \emph{per simulator}: a new structure requires new simulations and a newly trained network. 

HyperODE addresses both directions with the same shared model. Because the surrogate is differentiable, we can calibrate a system by gradient descent through the trained forward model, optimizing a parameter distribution whose induced trajectory bands are
calibrated against the observation. Alternatively, we proposed a single encoder with the trained decoder that maps a noisy trajectory directly to a parameter distribution in one forward pass, giving millisecond-scale inference across every structure with no per-system retraining.

We evaluate calibration with the weighted interval score (WIS), a proper scoring rule from the forecasting literature that jointly rewards sharpness and calibration \citep{bracher2021evaluating}, together with interval coverage. WIS is exactly the mean pinball (quantile) loss averaged over the predicted quantile levels, so minimizing it during training is a distribution-free way to fit calibrated bands -- connecting our objective to quantile regression \citep{koenker2001quantile} and conformal-style calibration \citep{shafer2008tutorial}.


\paragraph{Foundation models and learning ODEs.}
There have been attempts to create ``foundation models'' for ODEs -- \emph{single} model that, given an observed trajectory, discovers the dynamics of a \emph{new} system in one forward pass. These include ODEformer \citep{dascoli2024odeformer} that infers ODE symbolically, and Foundation Inference Models for ODEs (FIM-ODE) \citep{mauel2026fimode} that infers a neural field. However, they \emph{discover} the vector field from data, whereas HyperODE \emph{exploits} the known structure that is available in compartmental and reaction-network modeling. These foundation models are demonstrated only on low-dimensional systems (ODEformer and FIM-ODE are evaluated up to three to four state dimensions), and emit point estimates rather than calibrated uncertainty. Instead, HyperODE is not limited by scale (demonstrate for 500+ coupled states) and returns calibrated quantile bands. 

\paragraph{ODEs covered by HyperODE}
HyperODE targets mass-conserving systems with bounded states, and whose right-hand sides are polynomials, which spans compartmental epidemic models (SIR, SEIR, and their metapopulation variants) and a broad space of chemical reaction networks. However, two observations widen this scope well beyond what the definition suggests. First, although the training class is mass-conserving and autonomous, the learned operator extrapolates zero-shot to systems that violate these assumptions: systems with birth and death  rates, and externally forced, time-varying transmission. Second, a large family of ODEs, including those with rational, exponential, and trigonometric terms, can be rewritten as equivalent systems with only quadratic terms by introducing auxiliary variables using quadratization / quadratic recasting~\citep{kerner1981universal,carothers2005polynomial,hemery2020complexity}. A surrogate applicable to quadratic mass action may reach substantial portion of the ODEs that arise in practice.

\paragraph{Graph and hypergraph neural networks}
Message-passing graph neural networks aggregate information over pairwise edges and are effective learning processes over networks \citep{battaglia2016interaction,sanchezgonzalez2020learning,li2020fourier}. Hypergraph neural networks generalize message passing to higher-order relations, in which a single hyperedge connects an arbitrary set of nodes \citep{feng2019hypergraph}. We adopt a hypergraph representation because it allows us to naturally group multiple states appearing in a mass-action monomial, but our operator departs from generic (hyper)graph networks in one decisive respect: it carries an explicit multiplicative \emph{product} of its source states as an inductive bias, matching the mass-action law exactly rather than approximating it through learned message functions. Our ablation confirms that the hyperedge view matters -- clique-expanded GCN and GAT baselines, which lack it, are several times worse.


\section{Problem Setup and Metrics}

We first consider the class of compartmental ODEs $\dot X = f(X;\theta)$, where $f$ is a sum of polynomial (mass-action) terms and total mass is conserved. Concretely this includes meta-population $\mathrm{SIS}$, $\mathrm{SIR}$, $\mathrm{SI}$, $\mathrm{SIRS}$, and $\mathrm{SEIR}$ models over an arbitrary number $k$ of coupled groups, as well as randomly generated polynomial reaction networks. Each system is specified by a known \emph{structure} (which states appear in which terms) and a vector of rate parameters $\theta$. These parameters are described as \textit{distributions} rather than fixed values, thus introducing uncertainty.

We study two tasks. The \emph{forward} task takes a structure, an initial condition, and a distribution over $\theta$, and predicts per-state trajectory \emph{quantile bands} --the induced uncertainty on $X(t)$. The \emph{inverse} task takes a single noisy observed trajectory together with the known structure and recovers a \emph{distribution} over $\theta$ consistent with it in order to generate trajectory quantile bands over the observations.
Our data splits are out-of-distribution (OOD) by design, so that test performance measures genuine transfer rather than interpolation. Training uses $\mathrm{SIS}$/$\mathrm{SIR}$/$\mathrm{SI}$ families plus randomly generated ODEs (emulating reaction networks); validation uses the held-out $\mathrm{SIRS}$ family; and the test set uses the entirely unseen $\mathrm{SEIR}$ family plus random ODE structures unseen during training.


\section{Method: HyperODE}

\begin{figure}[!ht]
\includegraphics[width=\columnwidth]{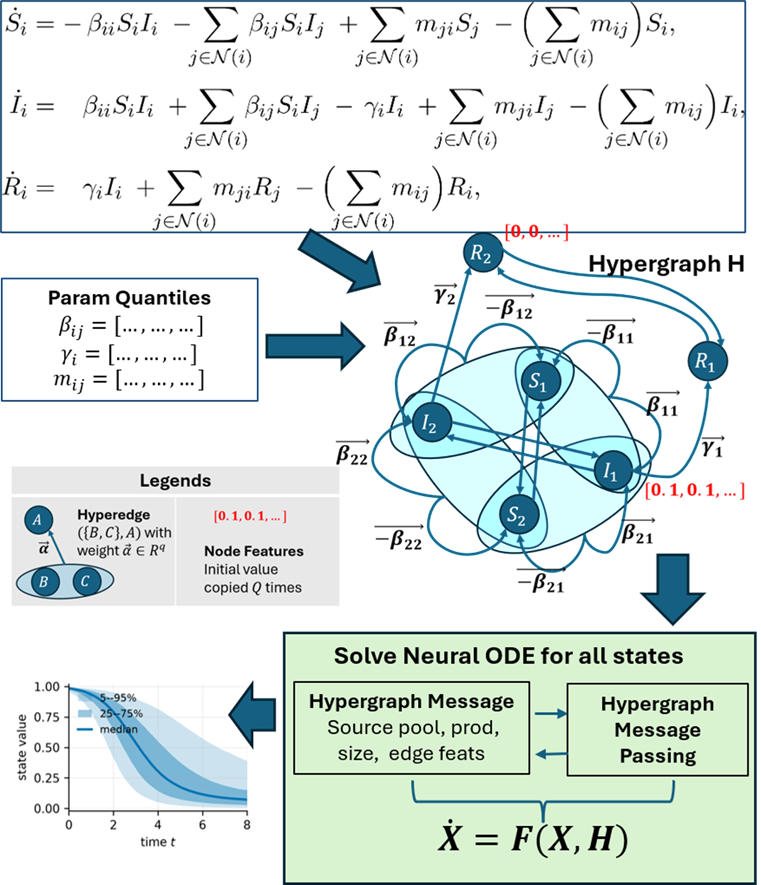}
\caption{Overview of HyperODE forward surrogate. It represents the given ODE as a hypergraph, takes uncertain parameters in the form of quantiles, and directly outputs trajectory distributions.}
\label{fig:overview}
\end{figure}

\paragraph{ODE as a directed hypergraph}
We represent an ODE structurally using a hypergraph (Figure~\ref{fig:overview}). Each state variable becomes a \emph{node}, and each additive monomial term on a right-hand side becomes a directed \emph{hyperedge} whose source set is the states involved in that term and whose target is the state it updates. For instance, a term $-\beta S_1 I_1$ contributing to $\dot S_1$ contributes a hyperedge with sources $\{S_1, I_1\}$. The coefficient $\beta$ is introduced in the structure as a $q$-dimensional weight vector on the hyperedge. The vector represents a vector of quantiles encoding the provided distribution. In this way, the representation is distribution-independent. The node feature is a $Q$-dimensional vector representing the quantiles of the state at time $t$. So, at $t=0$, the vectors, when the state has no uncertainty, all quantiles are the same -- the initial value.

Our objective is to train a hypergraph neural network $F$ on this hypergraph $H$, such that the solution to $\dot{\mathbf{X}} = F(\mathbf{X}, H)$ with the given initial values results in the $Q$ trajectories per state that represent the quantiles of the solutions to the original ODE $\dot{X} = f(X)$. Here $\mathbf{X} \in {R}^{n \times Q}$ represents the $Q$ quantiles of all the $n$ states in the system.

\paragraph{Shared hyperedge operator}

Each node $v$ carries a hidden state $h^{(l)}v$ that is refined over $L$ layers and initialized to the state value, $h^{(0)}v = x_v$. For a hyperedge $h$ with source set $S_h$, target $t_h$, and coefficient $c_h$, layer $l$ pools its sources through a shared MLP $\phi^{(l)}$, forms the mass-action product from the raw state, concatenates these with $|S_h|$ and edge features $e_h$, and emits a message through a second shared MLP $\psi^{(l)}$; each node then adds its incoming messages to a self-transform:

\[
\begin{aligned}
\mathrm{pool}^{(l)}_h &= \textstyle\sum_{s\in S_h}\phi^{(l)}\!\big(h^{(l)}_s\big), \quad
\mathrm{prod}_h       = \textstyle\prod_{s\in S_h} x_s ,\\
z^{(l)}_h             &= \big[\,\mathrm{pool}^{(l)}_h,\ \mathrm{prod}_h,\ |S_h|,\ e_h\,\big], \quad
m^{(l)}_h             = c_h\,\psi^{(l)}\!\big(z^{(l)}_h\big),\\
h^{(l+1)}_v           &= \sigma_l\!\Big(W^{(l)}_{\mathrm{self}}\,h^{(l)}_v
                         + \textstyle\sum_{h:\,t_h=v} m^{(l)}_h\Big),
\end{aligned}
\]
where $\sigma_l$ is sigmoid for all layers except the final one where it is replaced by identity so that $h^{(L)}v$ is node $v$'s time derivative. The resulting field is integrated with a fixed-step solver to obtain trajectory quantiles for each state.

The explicit product $\prod_{s\in S_h} x_s$ supplies the exact mass-action monomial as an inductive bias. The network only needs to learn ``corrections'' rather than rediscover the product. 
As in a typical GNN, the operator's weights are not state-specific. They encode how states impact each other in the system, and so can generalize to different systems and scales. Stacking multiple layers helps capture higher complexities (we hope to capture trajectory distributions, not just mean or median outcome) and implicit correlations across coefficients. This may arise if coefficients are formed by composition of parameters (e.g., $\beta$ and $\beta/\gamma$). 

\subsection{Quantile (UQ) Forward Surrogate}

Each node (state) carries a vector of $Q = 5$ quantiles ($\tau \in \{.05,.25,.5,.75,.95\}$)
rather than a scalar, and the edge features encode the parameter distribution through
coefficient quantiles, keeping the representation distribution-free. We train the surrogate
against the $K$ Monte-Carlo rollouts $\{y^{(k)}\}$ of the true ODE with a pinball (quantile)
loss and two light regularizers---a monotonicity penalty against quantile crossing and a
range penalty that keeps predictions in the valid state interval:
\[
  \mathcal{L} = \mathcal{L}_{\mathrm{pin}}
  + \lambda_{\mathrm{mono}}\,\mathcal{L}_{\mathrm{mono}}
  + \lambda_{\mathrm{rng}}\,\mathcal{L}_{\mathrm{rng}},
\]
\begin{align*}
\mathcal{L}_{\mathrm{pin}}  &= \operatorname*{mean}_{k,t,n,q}\
   \rho_{\tau_q}\!\big(y^{(k)}_{t,n}-\hat q_{t,n,q}\big),\\
\mathcal{L}_{\mathrm{mono}} &= \operatorname*{mean}_{t,n,q}\
   \big[\,\hat q_{t,n,q}-\hat q_{t,n,q+1}\,\big]_+,\\
\mathcal{L}_{\mathrm{rng}}  &= \operatorname*{mean}_{t,n,q}\
   \big[-\hat q_{t,n,q}\big]_+^2 + \big[\hat q_{t,n,q}- C\big]_+^2,
\end{align*}
where $\rho_\tau(u)=\max(\tau u,(\tau-1)u)$ is the pinball loss and $[\cdot]_+=\max(\cdot,0)$.
The monotonicity term discourages quantile crossing, and the range term keeps each predicted quantile within bounds $[0,C]$. The bound is introduced to avoid instability due to large prediction. However, ablations reveal that the impact of this term is negligible.

The loss scores the predicted band against the $K$ Monte Carlo rollouts. We note that each rollout is an unbiased sample of the population WIS. As a result, with large enough number of training \emph{systems} (ODE structure and parameter distribution combinations), the number of rollouts $K$ per system \textbf{does not need to be large} (we use $K{=}48$). The Monte-Carlo WIS objective is a two-stage (cluster) sample of the population score - systems are primary units, rollouts secondary. So its variance is $\sigma_b^2/M+\sigma_w^2/(MK)$ and the cost-optimal number of rollouts per system is $K^\star=\sqrt{\sigma_w^2/\sigma_b^2}$, the classical optimal allocation \citep{cochran1977sampling}. Because WIS is an average of per-rollout scores (linear in the inner samples), the empirical objective is unbiased for the population WIS at every $K$ and small $K$ suffices when systems are diverse. A detailed analysis is presented in the Appendix. This enables generating thousands of systems for \textit{training and evaluations} without a high cost of generating ``ground truth'' per system.

\subsection{Inverse: Noisy Data to Calibration}

\paragraph{Gradient-Based Calibration}
Since the forward surrogate is fully differentiable, we can calibrate a system given noisy state trajectories using gradient descent. Holding the surrogate weights frozen, we treat the parameter quantiles as free variables and minimize the WIS between the calibrated trajectory bands and noisy observed trajectories by gradient descent. This approach requires multiple epochs and can be slower than MCMC on small systems. However, it is comparatively, highly scalable - MCMC cost grows super-linearly with the number of parameters.

\paragraph{Single-pass Encoder}
To accelerate calibration we propose a ``single-pass'' encoder trained to find the calibration in a single forward pass. We train an encoder $E_\eta$ that maps a noisy observed trajectory directly to a \emph{distribution} over parameters. It is the \emph{transpose} of the forward operator: while the surrogate pushes parameters forward through the hypergraph to a trajectory, the encoder pulls the trajectory back through the same structure to the parameters. Given an observation $y \in R^{T\times n}$ (one time series per state), it runs in four stages. \emph{(i) Per-node GRU.} A GRU with weights shared across nodes embeds each state's noisy time series:
\[
  h_v = \mathrm{GRU}\big(y_{:,\,v}\big)\in R^{d},\qquad v=1,\dots,n .
\]
\emph{(ii) Nodes to edges.} For each hyperedge $h$ with sources $S_h$ and target $t_h$,
\[
  e_h = \phi_E\!\Big(\textstyle\frac{1}{|S_h|}\sum_{s\in S_h} h_s,\; h_{t_h},\; |S_h|\Big)\in R^{d},
\]
where $\phi_E$ is a shared MLP.
\emph{(iii) Edges to parameters.} We reuse the \emph{known} affine map between parameters
and coefficients, $c = c_0 + W\theta$ with $W\in R^{H\times P}$ fixed by the ODE structure:
using the column-normalized incidence $A_{hp} = |W_{hp}|/\sum_{h'}|W_{h'p}|$, each parameter
embedding pools its edges, $g_p = \sum_{h} A_{hp}\, e_h$.
\emph{(iv) Monotone quantile head.} With global context $\bar h = \frac1n\sum_v h_v$,
\[
  \hat\theta_{p,:} = \mathrm{cumsum}\Big(\mathrm{softplus}\big(\mathrm{MLP}([\,g_p \,\|\, \bar h\,])\big)\Big),
\]
so each parameter's predicted quantiles ($Q_{\mathrm{in}}=5$ levels) are positive and
non-decreasing by construction.
The encoder is trained \emph{through the frozen forward surrogate as a fixed decoder} $F$:
it predicts $\hat\theta = E_\eta(y)$, $F$ re-simulates a band, and the loss is the
reconstruction WIS against the observation,
\[
  \mathcal{L}(\eta) = \mathrm{WIS}\big(D(\hat\theta),\, y\big).
\]
Freezing $H$ is what prevents the usual autoencoder collapse -- the decoder cannot bend to
meet a degenerate encoding -- so the only way to lower the loss is to emit parameters that
actually reproduce the data. 
is a genuine zero-shot inverse.



\section{Experiments}

We intende to demonstrate that: (i) HyperODE surrogate accurately replicates MC at a much lower runtime; (ii) HyperODE can be used to calibrate from noisy data; (iii) HyperODE extends to a large class of systems and families not seen in training (larger systems, non mass-conserving, external forcing). All our code is publicly available~\footnote{\url{https://github.com/scc-usc/hyperODE}}.

\paragraph{Setup.}
We generate data from two sources. \textbf{Meta-population compartmental
families}: SIS, SIR, SI, SIRS, and SEIR models replicated over
$k$ groups coupled on a ring, where each group may interact with other groups. We keep the number of interactions as $O(k)$. The coupling induces transmission across groups/patches and linear inter-patch mobility ($m_{ij}$), and every right-hand-side term is thus a monomial; the aggregate infectious count $I_{\mathrm{tot}}=\sum_i I_i$ is tracked as an extra node. 
\textbf{Random reaction networks} : mass-conserving systems of linear reactions $a\!\to\!b$ (rate $\kappa x_a$) and quadratic catalysed reactions $a\!\to\!b$ (rate $\kappa x_a x_c$, $c\neq a$) on $n\in[2,6]$ compartments whose states start on the simplex, with an optional aggregate pseudo-node summing a random subset of derivatives; conservation and non-negativity hold by construction. 
The per-instance parameter uncertainty is itself randomized -- each rate independently draws its distribution family from \{log-normal, uniform, truncated-normal\} and then picks random distribution parameters -- so the surrogate can be trained across many \emph{shapes} of parameter uncertainty. We also generate random initial values for these systems. For each (ODE, parameter distribution, initial value) configurations, we generate $K = 48$ samples from the distribution and solve the ODE to form the targets (as noted earlier, $K$ need not be large when computing loss across large batches). Trajectories are integrated with a fixed step to $t_{\max}{=}8$ ($40$ steps, $\mathrm{d}t{=}0.2$). \textbf{Splits are structure-level out-of-distribution:} only $\{\text{SIS},\text{SIR},\text{SI}\}$ and random networks are seen in training, SIRS is held out for validation, and \textbf{SEIR for test}; the validation/test random networks use disjoint structure seeds, so at evaluation both the compartmental \emph{family} and the specific random \emph{topologies} are unseen. Models train on $k\in\{1,2, \dots, 64\}$ and are evaluated out to $k{=}128$. We use 4270, 1600, and 1420 configurations for training, validation, and testing, respectively.

\paragraph{Metrics}
We use WIS and Coverage as our main metrics. WIS (the weighted interval score) is the mean pinball loss of the predicted quantiles across the levels $\tau\in\{0.05,0.25,0.5,0.75,0.95\}$, scored against the Monte-Carlo samples. Coverage cov90 is the empirical fraction of the ground truth falling inside the central $5$-$95\%$ predictive band. A perfectly calibrated model attains the nominal $0.90$, and both over- and under-coverage are miscalibrations.

\subsection{Forward Surrogate}

\begin{figure}[t]\centering
  \begin{subfigure}{0.49\columnwidth}\centering
    \includegraphics[width=\linewidth]{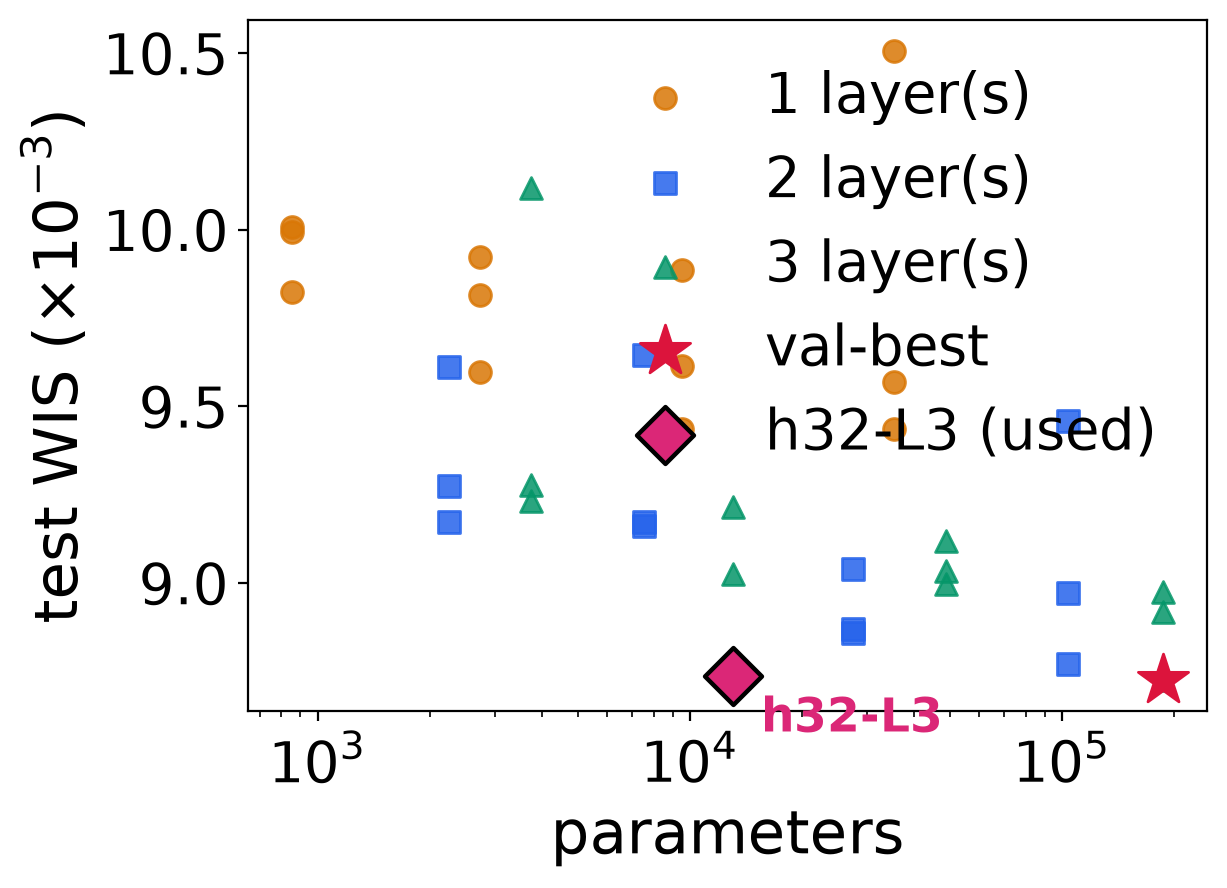}
    \caption{Test WIS vs.\ capacity.}
    \label{fig:forward}
  \end{subfigure}\hfill
  \begin{subfigure}{0.49\columnwidth}\centering
    \includegraphics[width=\linewidth]{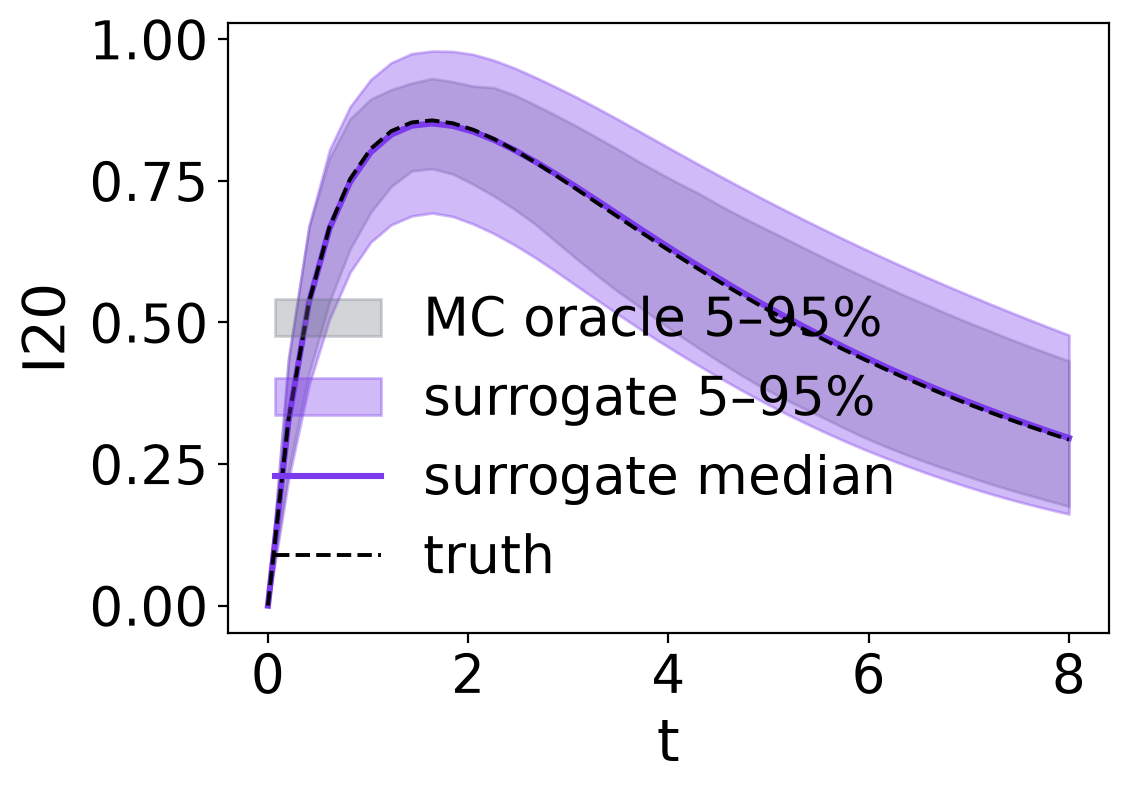}
    \caption{Forward predictive band.}
    \label{fig:band-fwd}
  \end{subfigure}
  \caption{\emph{Left}: Performance of all surrogate models and a sample output.}
  \label{fig:forward_full}
\end{figure}

We trained many forward surrogates with varying layers, dimensions per layer and learning rates. Based on the validation errors, we chose a 3 layer model with hidden dimension 32 as it achieved validation WIS close to the best model (3 layer, 128 dimension) with only around $10k$ parameters. The results for all the models on the test set are shown in Figure~\ref{fig:forward}.
The chosen h32-L3 model (${\approx}13$k parameters) attains a test WIS of $0.0087$ while remaining well-calibrated (cov90 $\approx 0.88$, close to the nominal $0.90$). We also observe that 2 and 3 layers significantly improve upon a single layer.

Figure~\ref{fig:band-fwd} shows the output of the surrogate compared to Monte Carlo oracle on a held-out large $\mathrm{SEIR}$ system. It is able to closely track the oracle and its distribution.

\begin{figure}[!ht]\centering
  \includegraphics[width=\columnwidth]{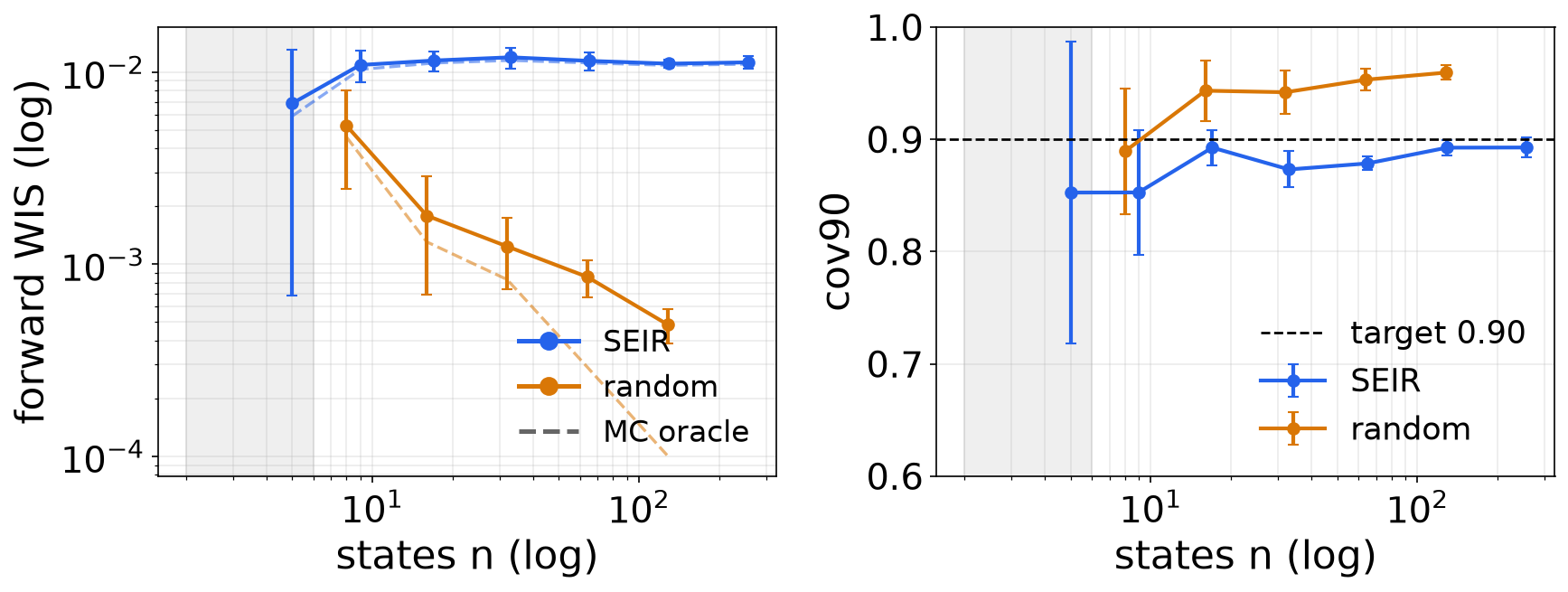}
  \caption{Zero-shot forward scale-generalization (both families, std error bars over structures).}
  \label{fig:scale}
\end{figure}

\paragraph{Baselines and ablations.}
We consider a \emph{baseline} which is a per-structure \emph{specialist}: a surrogate of the same architecture trained exclusively on one structure. Our general model beats the specialist by $7$--$11\%$ WIS at every $k$ even though $\mathrm{SEIR}$ is held out (Table~\ref{tab:specialist}). So cross-structure training actually helps the model rather than diluting it. We also ran variations of hyperGNN and GNN architectures. On the test set the \emph{hypergraph grouping} is decisive for raw accuracy (Table~\ref{tab:ablation}). \emph{GCN} and \emph{GAT} baselines are roughly $\gnnFac\times$ worse in WIS and attention does not help (GAT${\approx}$GCN), confirming that treating each monomial as a hyperedge matters. A more detailed ablation study for hypergraph features appears in Appendix. In particular, the product's value shows up out of distribution, at higher interaction order (see Figure~\ref{fig:scale}).

\input{tab_specialist}   
\input{tab_ablation}     

\begin{figure}[!ht]\centering
  \includegraphics[width=0.99\columnwidth]{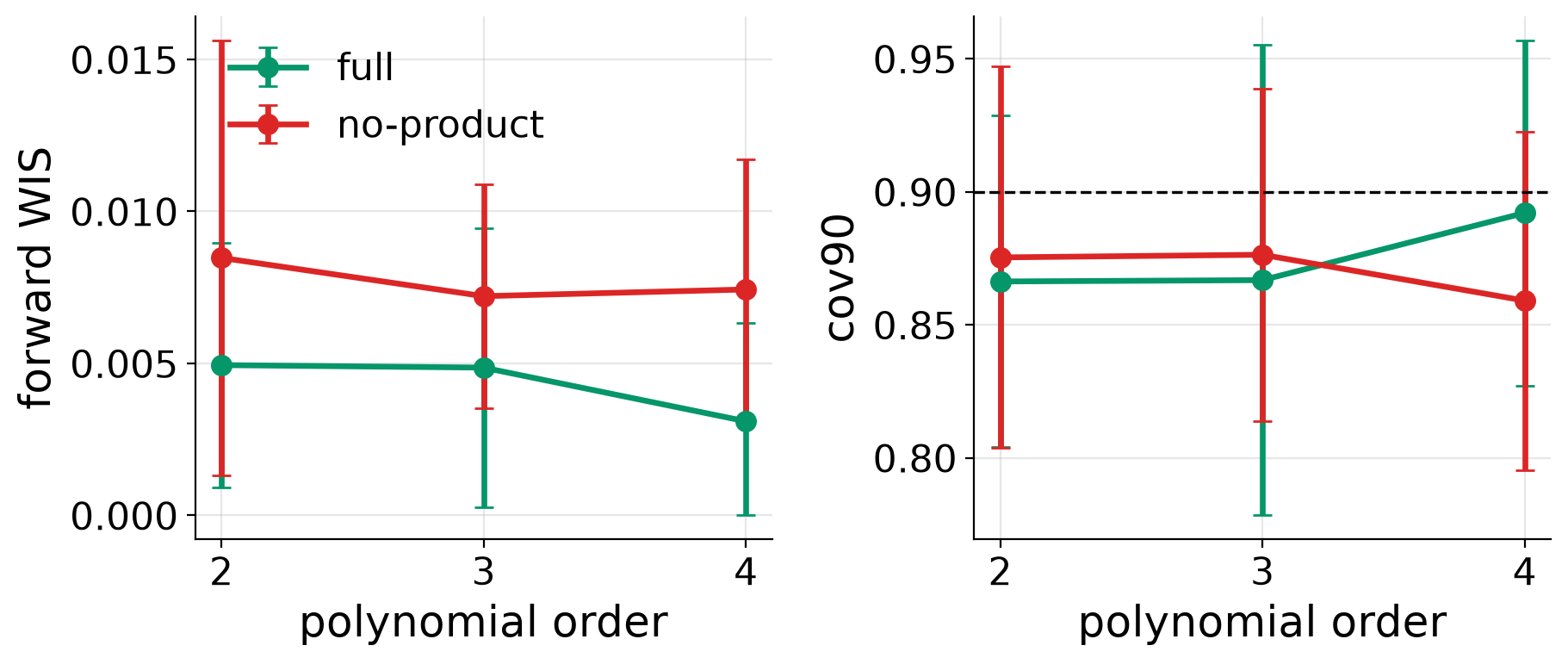}
  \caption{Zero-shot forward generalization to higher interaction order, highlighting the impact of prod feature.}
  \label{fig:highorder}
\end{figure}

\paragraph{Scaling and interaction order.}
We tested zero-shot generalization against a $K{=}200$-sample Monte Carlo oracle on held-out $\mathrm{SEIR}$ family up to $n{=}257$ states ($k{=}64$) and on random reaction networks forced far beyond their training size ($n{\le}6$) up to $n{=}128$. The MC oracle is obtained by scoring using $K/2$ samples against the rest $K/2$. With no retraining, the WIS and coverage deviation from $0.90$ remains low across scales (Figure~\ref{fig:scale}). Note that the standard deviation in the errors seem to decrease with size. This is due to the fact that larger systems have errors aggregated over larger number of states.
The forward surrogate also handles higher-order interactions it never saw: on random networks with cubic and quartic terms (order $2{\to}4$) the forward WIS and coverage stay essentially flat (Figure~\ref{fig:highorder}), and removing the product costs a modest but \emph{growing} penalty (${\approx}1.2\times$ at order $3$, $\hoFourFac\times$ at order $4$). This suggests that explicit product term helps with generalization to unseen orders.

\subsection{Inverse: Noisy Data to Calibration}

We sample one trajectory per configuration and introduce $5\%$ noise. On these, we perform calibrations three ways: a per-instance \emph{gradient} calibration ($300$ Adam steps through the frozen surrogate), our single shared \emph{single-pass encoder} (observation to parameters in one forward pass, no per-instance optimization), and \emph{MCMC} on the true ODE ($700$--$1600$ solves per problem). For training single pass encoder, we ran another hyperparameter sweep, and ended up with h128\_L2 as the forward surrogate decoder.

\begin{figure}[!ht]\centering
  \begin{subfigure}{0.49\columnwidth}\centering
    \includegraphics[width=\linewidth]{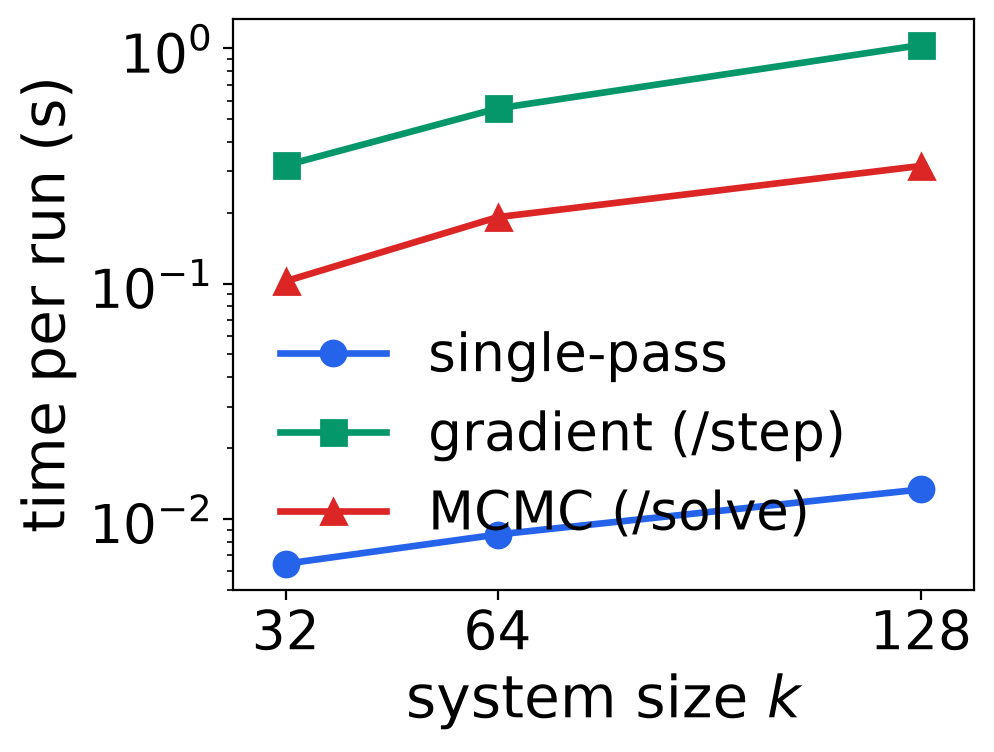}
    \caption{Inference cost \emph{per run} vs.\ system size $k$.}
    \label{fig:runtime}
  \end{subfigure}\hfill
  \begin{subfigure}{0.49\columnwidth}\centering
    \includegraphics[width=\linewidth]{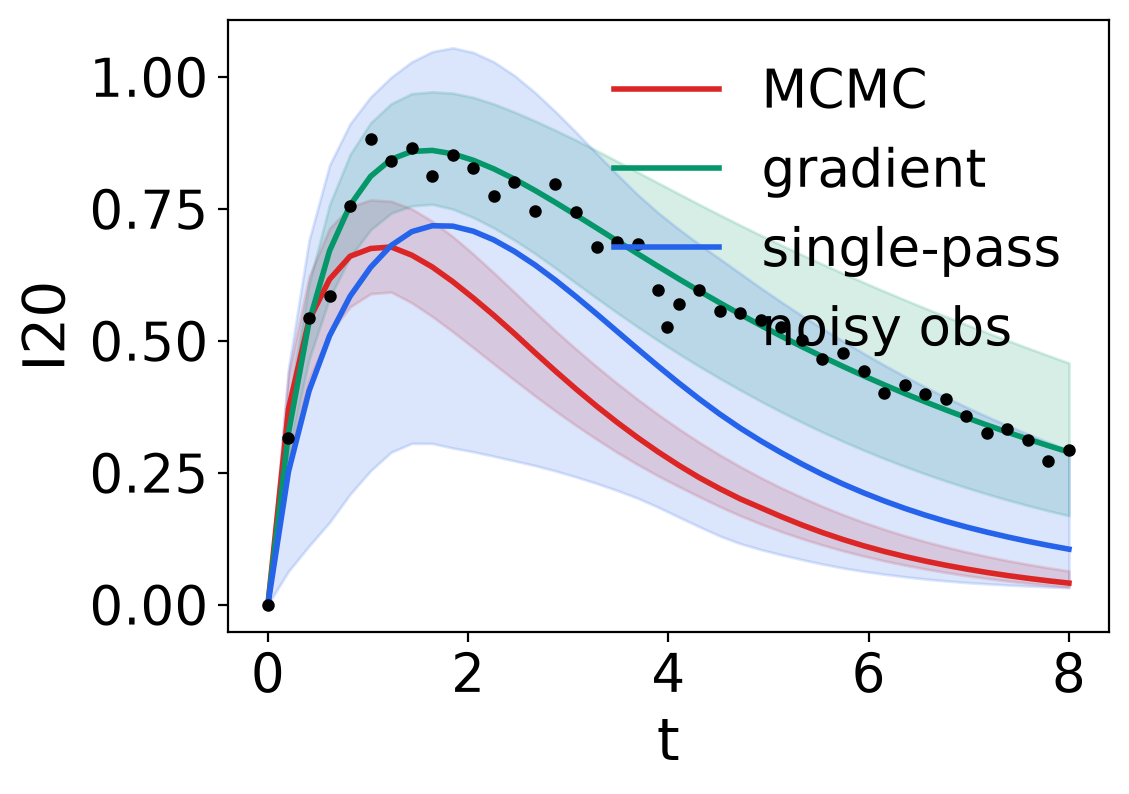}   
    \caption{Calibration bands, held-out large $\mathrm{SEIR}$ ($k{=}32$, $I_{20}$).}
    \label{fig:bands}
  \end{subfigure}
  \caption{Inverse calibration: runtimes and calibrated bands.}
  \label{fig:inverse-both}
\end{figure}

The single-pass encoder's decisive advantage is cost: since the MCMC iteration budget is a free parameter, we compare the intrinsic \emph{per-run} cost (Figure~\ref{fig:runtime}): one encoder pass ($6$--$13$\,ms) is ${\sim}15$--$25\times$ cheaper than a single MCMC solve and ${\sim}50$--$80\times$ cheaper than a single gradient step; accounting for iteration counts, the encoder is ${\sim}10^{3}$--$10^{4}\times$ faster end-to-end (milliseconds vs.\ $72$--$310$\,s). While MCMC may seem faster than our gradient-based approach, the number of iterations needed to achieve good results with MCMC scales super-linearly with number of states. This is reflected in the errors (Table~\ref{tab:oneshot}). The gradient-based inverse is the most accurate and could be accelerated by batching multiple systems leveraging the GPU. While both single-pass and limited-budget MCMC miscalibrate, MCMC performs worse, and achieving better results would require MCMC a much larger time budget. Figure~\ref{fig:bands} shows the calibration in one such instance.

Against per-structure simulation based inference -- NPE and NRE~\cite{papamakarios2016fast,hermans2020likelihood} with 2000 runs, our single encoder matches NPE on some models and outperforms both NPE and MCMC on the large $\mathrm{SEIR}$ system, with no per-structure training (Table~\ref{tab:npe}). We also compare against a foundation ODE model (FIM-ODE) in Table~\ref{tab:fim}. FIM-ODE is \emph{structure-blind} and infers the dynamics from data, whereas HyperODE is given the known structure and exploits it, so this comparison favors HyperODE by design and should be read as characterizing where each method applies rather than as a head-to-head on equal footing. FIM-ODE is further limited to systems with 3 states or fewer and produces point estimates rather than calibrated quantiles. Our single-shot encoder performs worse on less noisy data but overtakes FIM-ODE as noise increases, owing to its knowledge of the underlying ODE; the gradient-based approach outperforms both by a wide margin at while being a small model, at the cost of multiple gradient steps.

\input{tab_oneshot}
\input{tab_npe}   



\input{tab_fim}   

\subsection{Extensions}

\begin{figure}[t]\centering
  \includegraphics[width=\columnwidth]{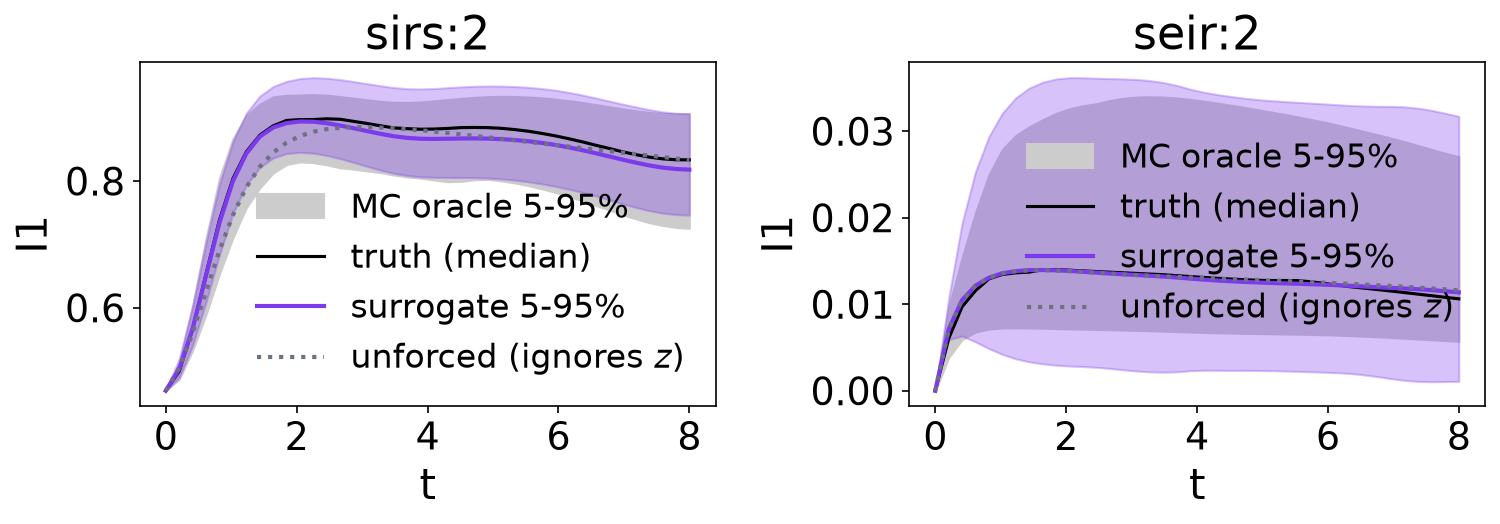}
  \caption{Introducing seasonal forcing via a dynamic graph.}
  \label{fig:forcing}
\end{figure}

\paragraph{External forcing.}
Since we have a neural ODE, we can swap a different hypergraph at each step. This allows for external  forcing such as seasonal transmission through a dynamic hypergraph. We can update the transmission edge with the known driver at each step $z(t)$. Doing this reproduces the forced $\mathrm{SEIR}$ dynamics to ${\approx}2\%$ relative error (rising to ${\approx}5\%$ on a larger instance) and yields a well-calibrated forced band ($\mathrm{cov90}\approx\forcingCov$), whereas an unforced control that ignores $z(t)$ instead produces the \emph{non-seasonal} trajectory (Figure~\ref{fig:forcing}).

\begin{figure}[!ht]\centering
  \includegraphics[width=0.99\columnwidth]{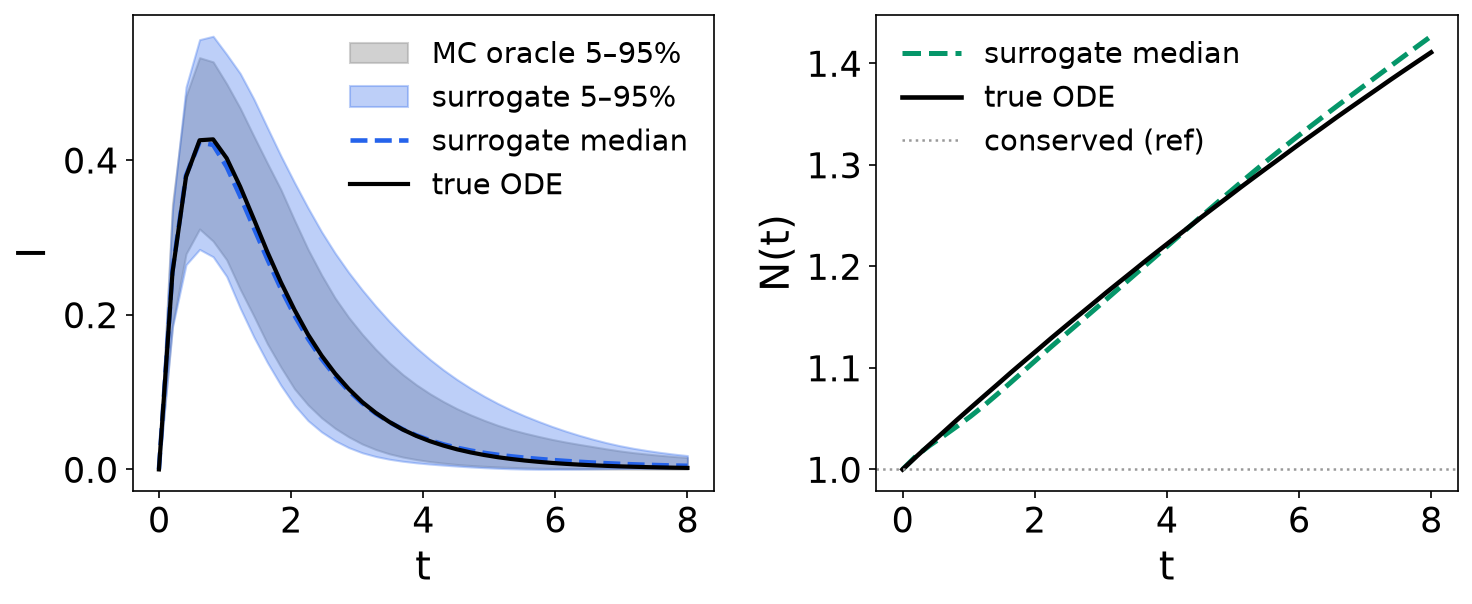}
  \caption{Zero-shot frequency-dependent transmission with non-polynomial terms $\beta S I/N$ via polynomialization.}
  \label{fig:freqdep}
\end{figure}

\paragraph{Vital dynamics (birth/death).}
The learned operator is \emph{local}---it sums monomial messages and never hard-codes a global conservation law -- so, reusing the dynamic-graph reformulation (clamped source/sink nodes for influx and removal), it extrapolates zero-shot to systems that violate conservation. Adding vital dynamics (birth rate $\Lambda$, death rate $\mu$) to $\mathrm{SEIR}$ makes the total population $N$ drift by up to ${\pm}41\%$, yet the frozen surrogate tracks the state trajectories to ${\approx}2\%$ relative error and $N$ itself to ${\approx}1\%$ (Figure~\ref{fig:freqdep}).

\paragraph{Handling non-polynomial terms.}
Frequency-dependent transmission $\beta S I / N$ is non-polynomial and thus outside the mass-action class. However, we can polynomialize by introducing an auxiliary state $M = 1/N$. This was used in Figure~\ref{fig:freqdep} mentioned above. This indicates that HyperODE applies to \emph{approximately} mass-conserving dynamics well beyond the class it was trained on.

\section{Conclusion}

We propose a single surrogate, HyperODE, that generalizes across the structure, scale, and interaction order of a class of approximately mass-conserving ODEs. It produces calibrated uncertainty in one forward pass and supports both a gradient and a single-pass inverse. By exposing the ODE as a directed hypergraph and applying a shared operator with an explicit mass-action product, it transfers zero-shot where conventional surrogates would require retraining. 

While we trained with only mass-conserving systems with maximum degree 2 terms, we have demonstrated that HyperODE applies to higher degrees, non-polynomials, and non-mass conserving systems as well. Further, external forces can be introduced into the system despite their absence in training. Our experiments are limited to compartmental models and reaction networks. A more diverse data generation can potentially make HyperODE more general and lead us towards a true foundation model. Finally, the performance gap between gradient-based inverse and single pass inverse suggests that the encoder architecture may be improved.

\newpage

\bibliography{aaai2027}

\newpage 
\input{appendix}


\end{document}

%% file: tab_specialist.tex
\begin{table}[t]
\centering\small
\setlength{\tabcolsep}{8pt}
\begin{tabular}{ccc}
\toprule
$k$ (SEIR) & generalist WIS $\downarrow$ & specialist WIS $\downarrow$ \\
\midrule
1 & $\mathbf{0.00762}$ & $0.00853$ \\
2 & $\mathbf{0.01033}$ & $0.01127$ \\
4 & $\mathbf{0.01069}$ & $0.01147$ \\
8 & $\mathbf{0.01056}$ & $0.01148$ \\
\bottomrule
\end{tabular}
\caption{One shared \emph{generalist} versus a \emph{specialist}
of the same architecture trained \emph{exclusively} on each held-out $\mathrm{SEIR}$-$k$.}
\label{tab:specialist}
\end{table}

%% file: tab_ablation.tex
\begin{table}[t]
\centering\small
\setlength{\tabcolsep}{6pt}
\begin{tabular}{lcc}
\toprule
model & WIS $\downarrow$ & cov90 (${\to}0.90$) \\
\midrule
hyperODE      & $0.0093${\scriptsize$\,\pm.0071$} & $0.872$ \\
GCN (vanilla)           & $0.0441${\scriptsize$\,\pm.0278$} & $0.868$ \\
GAT (attention)         & $0.0503${\scriptsize$\,\pm.0221$} & $0.914$ \\
\bottomrule
\end{tabular}
\caption{Ablation on the h32-L3 backbone for the forward surrogate.
(Figure~\ref{fig:highorder}).}
\label{tab:ablation}
\end{table}

%% file: tab_oneshot.tex
\begin{table}[t]
\centering\small
\setlength{\tabcolsep}{4pt}
\begin{tabular}{llccr}
\toprule
 & method & WIS $\downarrow$ & cov90 (${\to}0.90$) & runtime \\
\midrule
\multirow{3}{*}{small}
 & gradient & \textbf{.0053}\,{\scriptsize$\pm$.0027} & $0.97$ & 36\,s \\
 & encoder  & .0156\,{\scriptsize$\pm$.0121} & $0.88$ & \textbf{4.1\,ms} \\
 & MCMC     & .0066\,{\scriptsize$\pm$.0029} & $0.78$ & 4.1\,s \\
\midrule
\multirow{3}{*}{large}
 & gradient & \textbf{.0027}\,{\scriptsize$\pm$.0023} & $0.96$ & 191\,s \\
 & encoder  & .0129\,{\scriptsize$\pm$.0125} & $0.85$ & \textbf{9.5\,ms} \\
 & MCMC     & .0223\,{\scriptsize$\pm$.0236} & $0.37$ & 142\,s \\
\bottomrule
\end{tabular}
\caption{Inverse methods on noisy test data (cov90 target $0.90$).}
\label{tab:oneshot}
\end{table}

%% file: tab_npe.tex
\begin{table}[t]
\centering\small
\setlength{\tabcolsep}{6pt}
\begin{tabular}{llccc}
\toprule
 & method & rand & rand\,$k{=}32$ & SEIR\,$k{=}32$ \\
\midrule
\multirow{3}{*}{\shortstack[l]{ours\\(one model)}}
 & single-pass & $0.0067$ & $0.0018$ & $0.0227$ \\
 & gradient    & $\mathbf{0.0038}$ & $\mathbf{0.0009}$ & $\mathbf{0.0047}$ \\
\midrule
 classic & MCMC        & $0.0050$ & $0.0011$ & $0.0335$ \\
\midrule
\multirow{2}{*}{\shortstack[l]{per-struct.\\SBI}}
 & NPE & $0.0065$ & $0.0017$ & $0.0260$ \\
 & NRE & $0.0120$ & --       & --       \\
\bottomrule
\end{tabular}
\caption{Comparison against per structure simulation-based inference (SBI).}
\label{tab:npe}
\end{table}

%% file: tab_fim.tex
\begin{table}[t]
\centering\small
\setlength{\tabcolsep}{6pt}
\begin{tabular}{lccc}
\toprule
 & \multicolumn{3}{c}{fit MAE\,$\downarrow$ ($D{\le}3$ slice)} \\
\cmidrule(lr){2-4}
method & 5\% noise & 10\% & 25\% \\
\midrule
FIM-ODE (pretrained, 13M)   & $0.0203$ & $0.0256$ & $0.0623$ \\
HyperODE (gradient, 13k)         & $0.0044$ & 0.0069       & 0.0157       \\
HyperODE (one-shot, 140k)         & $0.0591$ & $0.0593$ & $0.0591$ \\
\bottomrule
\end{tabular}
\caption{FIM-ODE vs HyperODE for calibration from noisy data. FIM degrades with noise.}
\label{tab:fim}
\end{table}

%% file: appendix.tex
\appendix
\section{Monte-Carlo Sample Size Sufficiency}
Every system is scored against $K{=}48$ true-ODE rollouts during training and most of testing. Because WIS is an \emph{average}
of per-rollout scores, the empirical objective
$\hat R_{M,K}=\frac1M\sum_m\frac1K\sum_k \ell(f;g_m,y_{m,k})$ is a two-stage (cluster) sample of
the population score (systems are the primary units, rollouts the secondary), so it is
unbiased for the population WIS at \emph{every} $K$, with the classical two-stage
variance~\citep{cochran1977sampling}
\[
  \operatorname{Var}[\hat R_{M,K}]=\frac{\sigma_b^2}{M}+\frac{\sigma_w^2}{MK},
\]
where $\sigma_b^2$ is the between-system and $\sigma_w^2$ the within-system variance of the
per-rollout WIS. Averaging over the inner rollouts is what removes any small-$K$ bias. Two consequences follow. At a fixed budget $N{=}MK$ the
variance is $\tfrac{K}{N}\sigma_b^2+\tfrac1N\sigma_w^2$, minimized by spending on more systems
(small $K$, large $M$); and extra rollouts stop helping once $\sigma_w^2/(MK)$ is dominated by
$\sigma_b^2/M$, i.e.\ for $K\gtrsim\rho\equiv\sigma_w^2/\sigma_b^2$.

\paragraph{How far is the reported WIS from the true loss?} The train/test WIS
standard error due to $K$ rollouts for each of the $M$ systems is given by
\[
  \operatorname{SE}[\hat R_{M,K}]=\sqrt{\frac{\sigma_b^2}{M}+\frac{\sigma_w^2}{MK}}.
\]
We estimate $\sigma_w^2,\sigma_b^2$ directly from the generated data by a one-way random-effects
ANOVA: for each system $m$, the MC-oracle band (empirical quantiles of its $K$ rollouts) is scored
against each rollout $k$, giving per-rollout scores $\ell_{m,k}$; then
\[
\hat\sigma_w^2=\tfrac1M\sum_m \mathrm{Var}_k(\ell_{m,k}),\qquad
\hat\sigma_b^2=\mathrm{Var}_m(\bar\ell_m)-\hat\sigma_w^2/K .
\]
The components $\sigma_w^2,\sigma_b^2$ are themselves estimated, but from the full pool ($M{=}4270$
configurations at $K{=}48$ each), so $\sigma_w^2$ carries $M(K{-}1)\approx2{\times}10^5$ and
$\sigma_b^2$ carries $M{-}1$ degrees of freedom. Table~\ref{tab:kstar}, shows the resulting standard error
from the choice of K. Observe that increasing K would have limited impact on the computed loss.

\begin{table}[!ht]\centering
\setlength{\tabcolsep}{6pt}
\begin{tabular}{lcccc}
\toprule
 & \multicolumn{2}{c}{variance ($10^{-5}$)} & \multicolumn{2}{c}{$\operatorname{SE}[\hat R]$ ($10^{-4}$)} \\
\cmidrule(lr){2-3}\cmidrule(lr){4-5}
scope & $\sigma_w^2$ & $\sigma_b^2$ & $K{=}48$ & $K{\to}\infty$ \\
\midrule
train & $5.2$ & $7.4$ & $1.33$ & $1.32$ \\
test  & $4.3$ & $4.0$ & $1.70$ & $1.68$ \\
\bottomrule
\end{tabular}
\caption{Sampling accuracy of the reported (set-averaged) WIS.
$\operatorname{SE}[\hat R_{K,M}]=\sqrt{\sigma_b^2/M+\sigma_w^2/(MK)}$, with $M{=}4270$ (train) and
$1420$ (test). The $K{=}48$ and $K\!\to\!\infty$ columns differ by ${<}1.2\%$.}
\label{tab:kstar}
\end{table}

\section{Inverse: Batched-Inference Throughput}
Table~\ref{tab:batchtime} measures per-problem wall time
vs.\ batch size $B$ on a single GPU. The \emph{gradient} inverse benefits most -- batching fills the GPU
across its $300$ optimization steps, cutting per-problem time by an order of magnitude. This
substantiates the main-text remark that the gradient inverse can be accelerated by batching.

\input{tab_batchtime}

\section{Extension: Generalized Product}\label{sec:genprod}
The most important extension of the operator is that its mass-action product feature is not locked
to a monomial: swapping it at inference lets one pretrained model integrate systems far outside the
training class. We first establish that \emph{within} the mass-conserving training class the product
feature does not matter at all, so its value is entirely a generalization effect.

\subsection{In distribution, the product feature is redundant}\label{sec:diff}
\begin{figure*}[t]\centering
  \begin{subfigure}{0.35\textwidth}\centering
    \includegraphics[width=\linewidth]{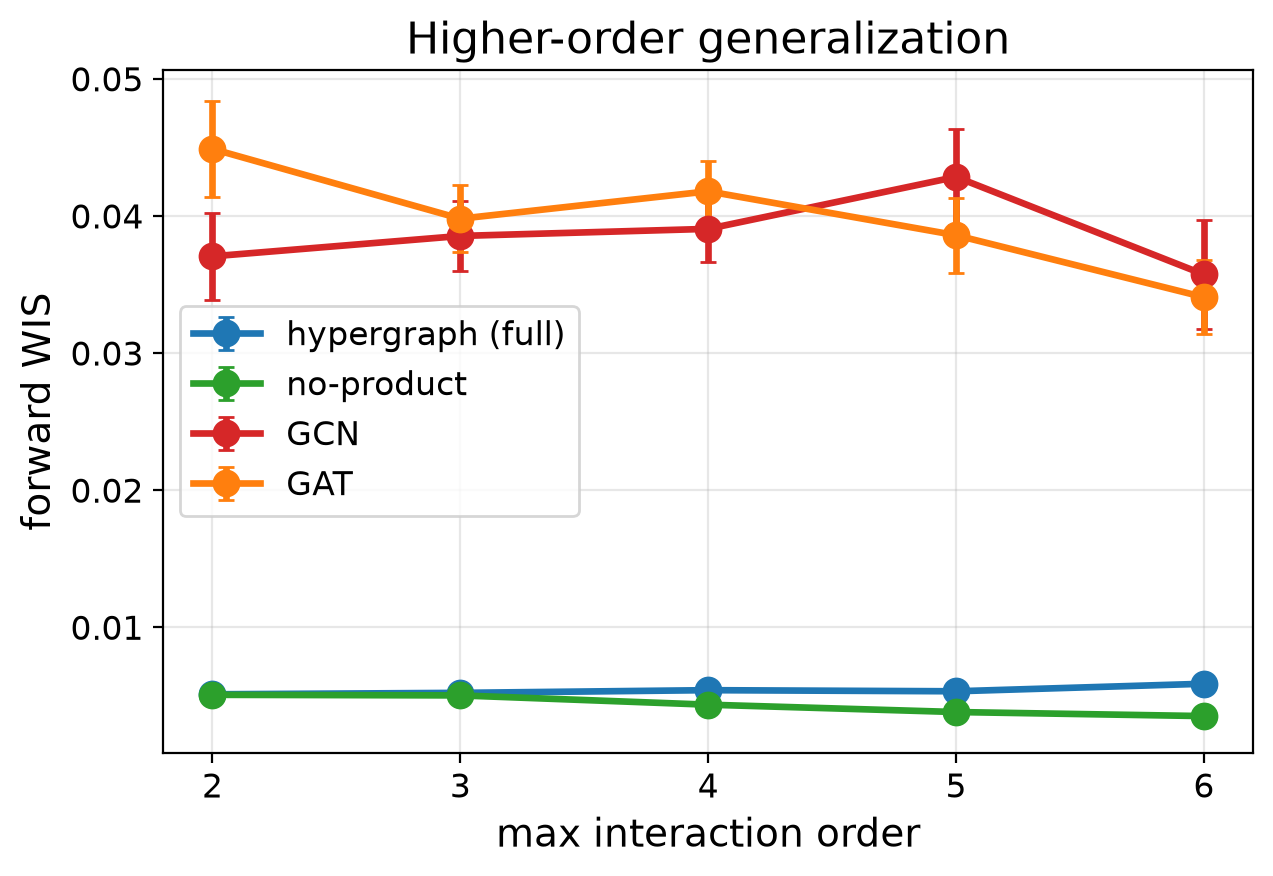}
    \caption{Interaction order.}
  \end{subfigure}\hfill
  \begin{subfigure}{0.6\textwidth}\centering
    \includegraphics[width=\linewidth]{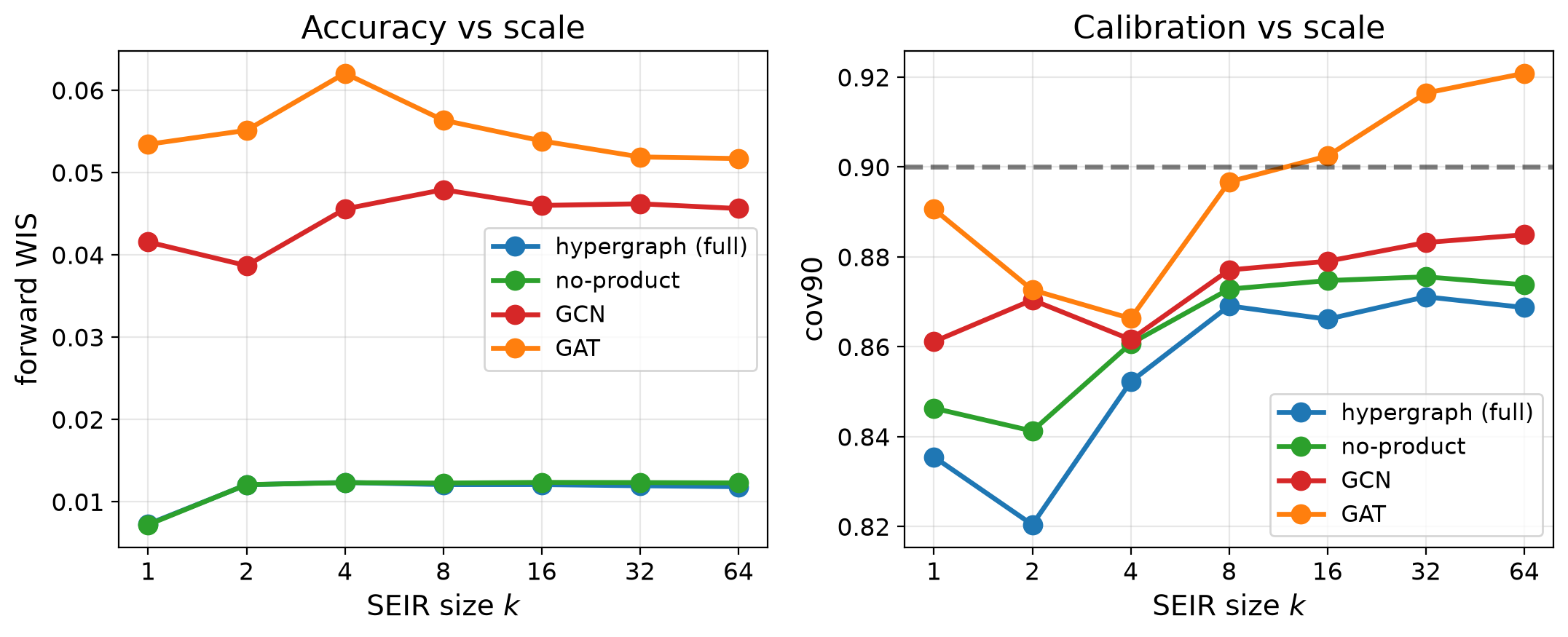}
    \caption{System scale (SEIR $k$).}
  \end{subfigure}
  \caption{Operator differentiation at h32-L3: interaction order (left) and system scale (right). 
  The choice of explicit product term does not matter within our test set, but it does in zero-shot 
  generalization to other functions (see Table~\ref{tab:nonpoly}).}
  \label{fig:diff}
\end{figure*}

We compare our approach with inclusion (full vs  no-product) and exclusion of the product feature across
\emph{interaction order} ($2\to6$, random reaction networks) and the \emph{system scale}
(held-out SEIR, $k{=}1\to64$, $n{=}5\to257$) in Figure~\ref{fig:diff}. The large, robust effect is
\emph{hyperedge grouping}: the GCN/GAT clique-expansions are $7$--$8\times$ worse on the order axis
and $4$--$5\times$ worse on the scale axis, at every order and every scale. 
For some other backbones, we found that the pooling term was necessary to prevent the training from diverging.
Generally, we observe that for mass-conserving training class, explicit computation of product does not matter. 
However, we show next that such explicit computation is necessary for zero shot generalization across products.

\input{tab_nonpoly}

\begin{figure*}[!ht]\centering
  \includegraphics[width=\textwidth]{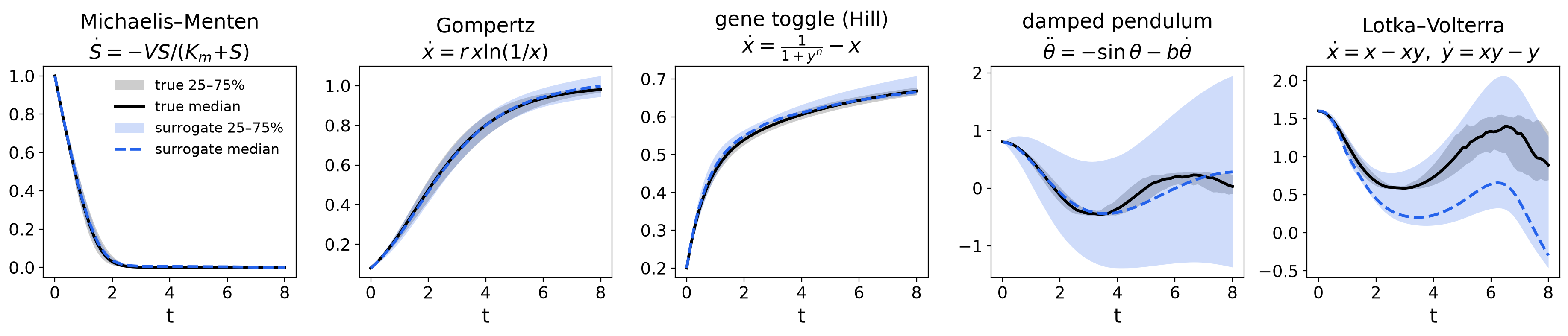}
  \caption{Generalized product across ODE families (zero-shot, one pretrained operator): damped pendulum, Michaelis-Menten, Gompertz, Hill toggle, and a Lotka-Volterra control.}
  \label{fig:genprodfam}
\end{figure*}

\subsection{Out of distribution, the product becomes essential}
The mass-action product $\mathrm{prod}_h=\prod_{s\in S_h}x_s$ is a per-hyperedge \emph{feature}
computed from the raw sources at every solver substep. Nothing ties it to a monomial product -- it can be
replaced at inference by \emph{any} driving nonlinearity $g_h(x_{S_h})$. Feeding the correct nonlinearity lets the same operator integrate systems outside the mass-action class. In addition to swapping, we also ensure the following. (i) Every state is shifted and normalized into $[0,1]$. (ii) The mass-action product $\prod_s x_s$ is always \emph{non-negative}, since $\psi$ only ever saw non-negative $\mathrm{prod}$.

\paragraph{No-prod with polynomialization.} To assist the \emph{no-prod} version address this we need to polynomialize as discussed earlier, as it was only trained to approximate products. Many of these terms \emph{can} be polynomialized through introduction of auxiliary variables. However, this may also lead to states going unbounded (e.g., $X=1/Y$, when $Y$ is small) or lead to large coefficients. Large coefficients can be reduced by changing the timescale, however, this may cause changes in step size or time horizon that are difficult for the surrogate.

We swapped the transmission
edge's product, $\prod_s x_s\to g(S,I)$ to the following functions forms -- sub/super-linear $SI^{p}$, 
saturating (Holling-II) $SI/(1{+}\alpha I)$, Ricker $SI\,e^{-\alpha I}$, and Hill $SI^{n}/(K^{n}{+}I^{n})$.
For all these cases, HyperODE with \emph{swap} with no retraining continues to demonstrate good WIS and coverage.
\emph{no-prod} only manages to keep up when the functions are approximately multiplicative for small $I$ (saturating and Ricker). Polynomialization helps, however it remains worse than \emph{swap} and fails when time-scale ($\lambda$) needs to be changed significantly.

\subsection{Other Families}
Figure~\ref{fig:genprodfam} applies the generalized product zero-shot across five ODE families from different domains, using the single pretrained operator with no retraining. Each system contains a term outside the mass-action class; at inference we substitute that hyperedge's product feature $\prod_{s}x_s$ with the term's actual driving nonlinearity $g(\cdot)$, after shifting/normalizing the states into $[0,1]$ and keeping $g$ non-negative. On the dissipative/monotone families (Michaelis--Menten, Gompertz, Hill toggle) the surrogate's $25$--$75\%$ predictive band closely matches the Monte-Carlo band under parameter uncertainty. On the oscillators (pendulum, Lotka--Volterra) the median tracks the true trajectory trend approximately zero-shot while the band over-covers.

\section{Experimental Setup}

\subsection{Architecture and Training}
The hyperedge operator uses two MLPs $\phi,\psi$: each hyperedge $h$ emits a message
$c_h\,\psi([\sum_s\phi(x_s),\ \prod_s x_s,\ |S_h|,\ \mathrm{efeat}_h])$, with the product computed
in log-space for numerical stability and always taken from the raw state values. The forward
sweep spans hidden width $\in\{16,32,64,128\}$, learning rate
$\in\{5{\times}10^{-4},10^{-3},5{\times}10^{-3}\}$, and depth $\in\{1,2,3\}$; every model uses an
Euler solver with 2 substeps, batch size 32, and no range penalty, and is selected by validation
pinball loss. The selected model is h32-L3 ($\approx$13k parameters).

Training minimizes the pinball loss plus a monotonicity penalty ($\lambda{=}1$) with Adam and a
plateau-based learning-rate schedule, gradient clipping at 5, skipping any batch with a non-finite
loss, and best-by-validation checkpointing. The single-pass inverse encoder maps observations
through a per-node GRU, a hyperedge message-passing layer, and an affine map $W$ to parameters,
followed by a monotone quantile head.

\paragraph{Hardware}
All neural-network training and inference ran on a single NVIDIA V100 GPU (32\,GB).
Data generation and the classical
baselines are CPU-bound and were run on 28-core Intel Xeon with 256 GB RAM.
The batched-inference timings (Table~\ref{tab:batchtime})
were measured on the same V100.

\subsection{Data Generation}
Two family generators supply structures. Meta-population compartmental models (SIS/SIR/SI/SIRS/SEIR)
use $k$ groups with ring coupling, mass-action transmission, and linear mobility, while random
reaction networks place mass-conserving linear and quadratic reactions on $n\in[2,6]$ compartments,
with an optional aggregate pseudo-node. A test-only higher-order generator adds order-$m$ reactions
$a\to b$ at rate $k\prod_{s\in S}x_s$ over $m$ distinct compartments including $a$, preserving
conservation and non-negativity; order $\ge3$ never appears in training.

Splits are out-of-distribution by both family and seed: training draws SIS/SIR/SI plus random
networks, validation the held-out SIRS family plus disjoint random networks, and testing the
entirely unseen SEIR family plus further disjoint random networks --- $4270$, $1600$, and $1420$
configurations respectively. For forward UQ, each instance is solved with $K{=}48$ true-ODE rollouts
to $t_{\max}{=}8$ over $T{=}40$ steps.
The encoder trains against the frozen surrogate decoder with
observation noise drawn log-uniformly in $[1\%,25\%]$. All experiments use fixed random seeds, and
the data splits are disjoint by seed and by family.

\subsection{Metrics}
WIS is the mean \emph{pinball} (quantile) loss over the quantile levels
$\tau\in Q=\{.05,.25,.5,.75,.95\}$. For a predicted $\tau$-quantile $\hat q_\tau$ and an observation
$y$, the pinball loss
\[
  \rho_\tau(y,\hat q_\tau)
  =\begin{cases}
     \tau\,(y-\hat q_\tau), & y\ge\hat q_\tau,\\[2pt]
     (1-\tau)\,(\hat q_\tau-y), & y<\hat q_\tau,
   \end{cases}
\]
penalizes under-prediction with weight $\tau$ and over-prediction with weight $1-\tau$, and is
minimized in expectation by the true $\tau$-quantile. The score is averaged over all quantile levels
and evaluation points (time steps and states),
\[
  \mathrm{WIS}=\frac1{|Q|}\sum_{\tau\in Q}\rho_\tau(y,\hat q_\tau).
\]
The coverage at level $q$ is $\hat\tau_q=\Pr[Y\le\hat q_{\tau_q}]$, the $90\%$-band coverage is
cov90 $=\hat\tau_{.95}-\hat\tau_{.05}$ (target $0.90$). For the inverse
task we report the reconstruction WIS of the posterior-predictive band and coverage based on observed noisy trajectory.

\section{Additional Experiments}

\subsection{Forward Calibration Across Quantiles}
\begin{figure}[t]\centering
  \includegraphics[width=\columnwidth]{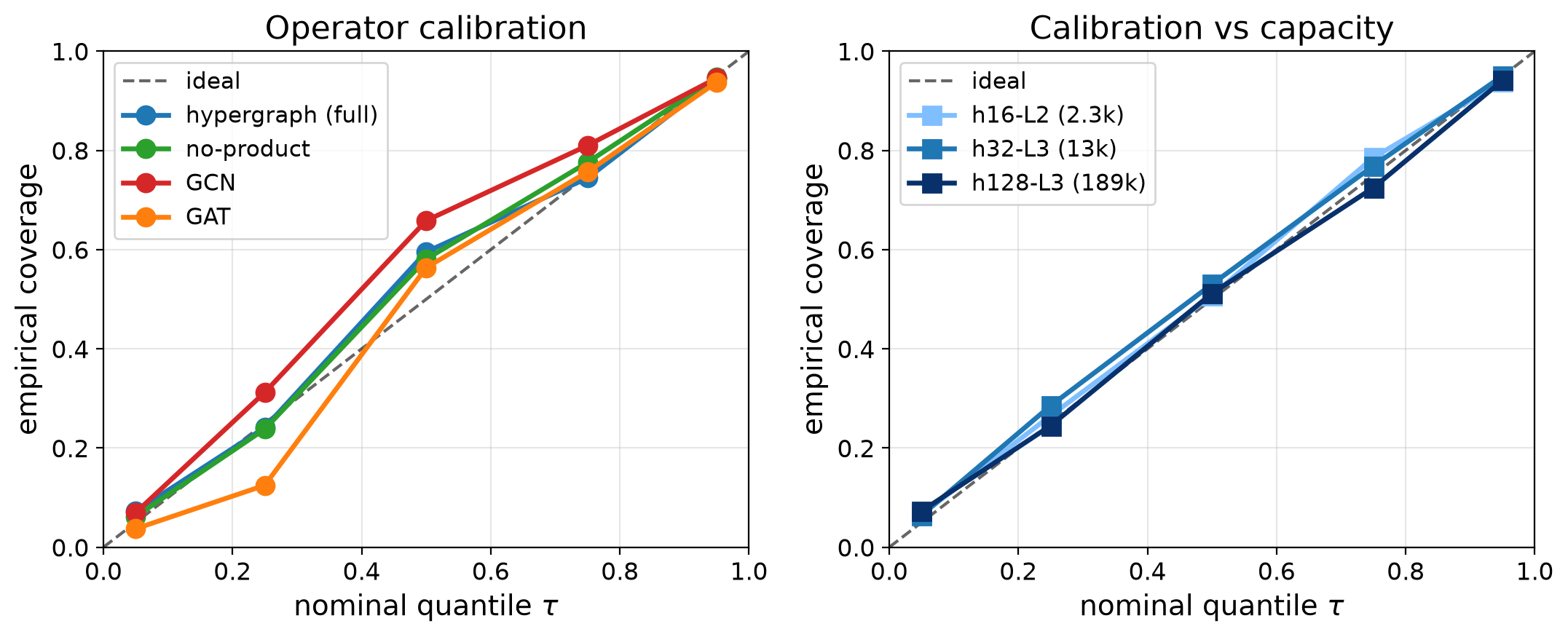}
  \caption{Reliability diagrams (empirical vs.\ nominal coverage; diagonal is perfect calibration).
  \emph{Left:} operator comparison. \emph{Right:} across model capacity.}
  \label{fig:reliability}
\end{figure}

Throughout the paper, for coverage we have used coverage at $0.90$ quantile (cov90). Figure~\ref{fig:reliability} gives the full
coverage diagram -- empirical vs.\ nominal coverage at every quantile level, the diagonal
being ideal. \emph{Left:} the hypergraph operator (full and no-product) tracks the diagonal
closely, whereas the GCN/GAT clique-expansions mis-place the interior quantiles (GCN's nominal
$25$--$75\%$ interval over-covers; GAT under-covers the lower tail). \emph{Right:} calibration is
essentially invariant to model capacity across the $80\times$ span (h16-L2 $\to$ h128-L3).
Aggregated over in-distribution random networks (order 2) and the held-out SEIR family.

\subsection{Forward Baselines: Floor, Oracle, and Specialists}
Table~\ref{tab:floor} places the surrogate between a naive floor and the MC oracle
ceiling. The MC oracle is the empirical band of the $K$ true-ODE solves, scored on a held-out $K/2$. Recall that the reported mean is across a large number of systems, making the size $K/2 = 24$ less relevant, given the variance across systems. 
The surrogate sits within $2.6\%$ of this floor in mean WIS and is marginally better on
the median (it avoids the oracle's finite-sample quantile noise), while beating the flatline
floor by $\sim14\times$.

\begin{table}[t]\centering\small
\begin{tabular}{lcc}
\toprule
estimator & WIS (mean$\pm$std) & median \\
\midrule
flatline ($x{\equiv}x_0$)  & $0.1248\pm.0736$ & $0.1344$ \\
MC oracle (irred.\ floor)  & $0.00867\pm.0062$ & $0.00953$ \\
surrogate h32-L3           & $0.00890\pm.0068$ & $\mathbf{0.00945}$ \\
\bottomrule
\end{tabular}
\caption{Forward baselines on the $1420$-structure test split (per-structure WIS): the surrogate
sits at the irreducible MC-oracle floor and $\sim14\times$ below the flatline floor.}
\label{tab:floor}
\end{table}

\subsection{Median of HyperODE vs ODE with median parameters}
Does the surrogate learn anything a single ODE solve does not? Specifically, we test if the surrogate median is just learning what a single ODE solve with median of parameters (\emph{solve-median}) would produce. On the test dataset \emph{solve-median} recovers the MC median to $\approx2.4\%$ relative error. So on the point estimate at moderate uncertainty a surrogate median does not earn a speedup, however, it does contribute the calibrated band and the differentiable inverse.

\begin{figure}[t]\centering
  \includegraphics[width=0.92\columnwidth]{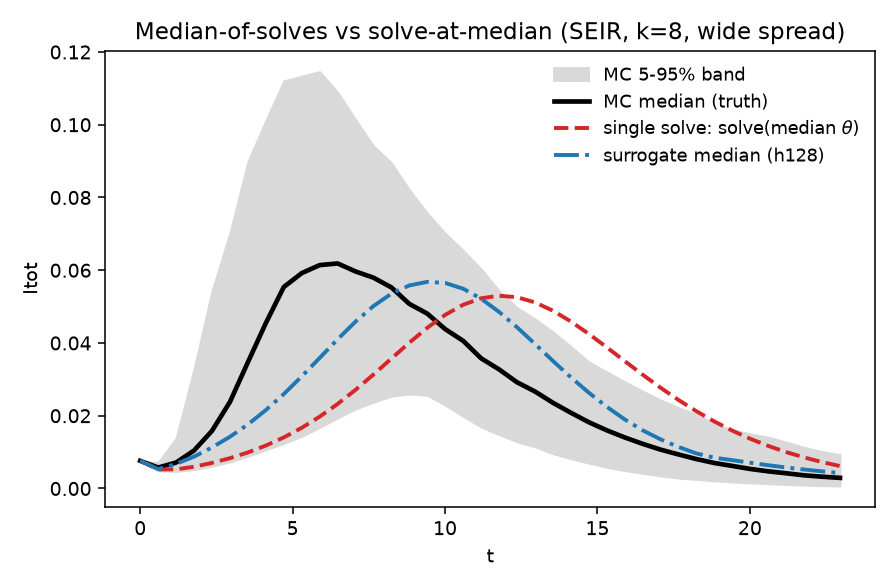}
  \caption{Median-of-solves vs.\ solve-at-median on a wide-spread SEIR ($k{=}8$). The single central
  solve mistimes the peak ($t\approx12$); the surrogate median tracks the true early rise
  ($t\approx6$--$9$), inside the MC band.}
  \label{fig:jensen}
\end{figure}

But that tie is specific to limited spread in parameters in the test set. We generated more test instances where the parameter uncertainty is around thresholds leading to significantly different peak heights and timings per run. 
Testing the pretrained h128 model (no retraining) on held-out SEIR across system size $k$ and
transmission spread $\mathrm{sig}$, the surrogate median beats a single solve on
$72$--$98\%$ of systems, growing with both size and uncertainty (Table~\ref{tab:jensen}), with
lopsided margins ($k{\ge}8,\mathrm{sig}{=}0.8$: wins by ${\sim}{+}0.05$, loses by ${\sim}{-}0.01$).
Figure~\ref{fig:jensen} shows one of the difficult test samples -- \emph{solve-median} produces a trajectory that even goes out of the 95\% interval. HyperODE forward surrogate, while not a perfect match, still remains within the interval.

\input{tab_jensen}

\subsection{Temporal and Step-Size Generalization}
\begin{figure}[!ht]\centering
  \begin{subfigure}{0.4\textwidth}\centering
    \includegraphics[width=\linewidth]{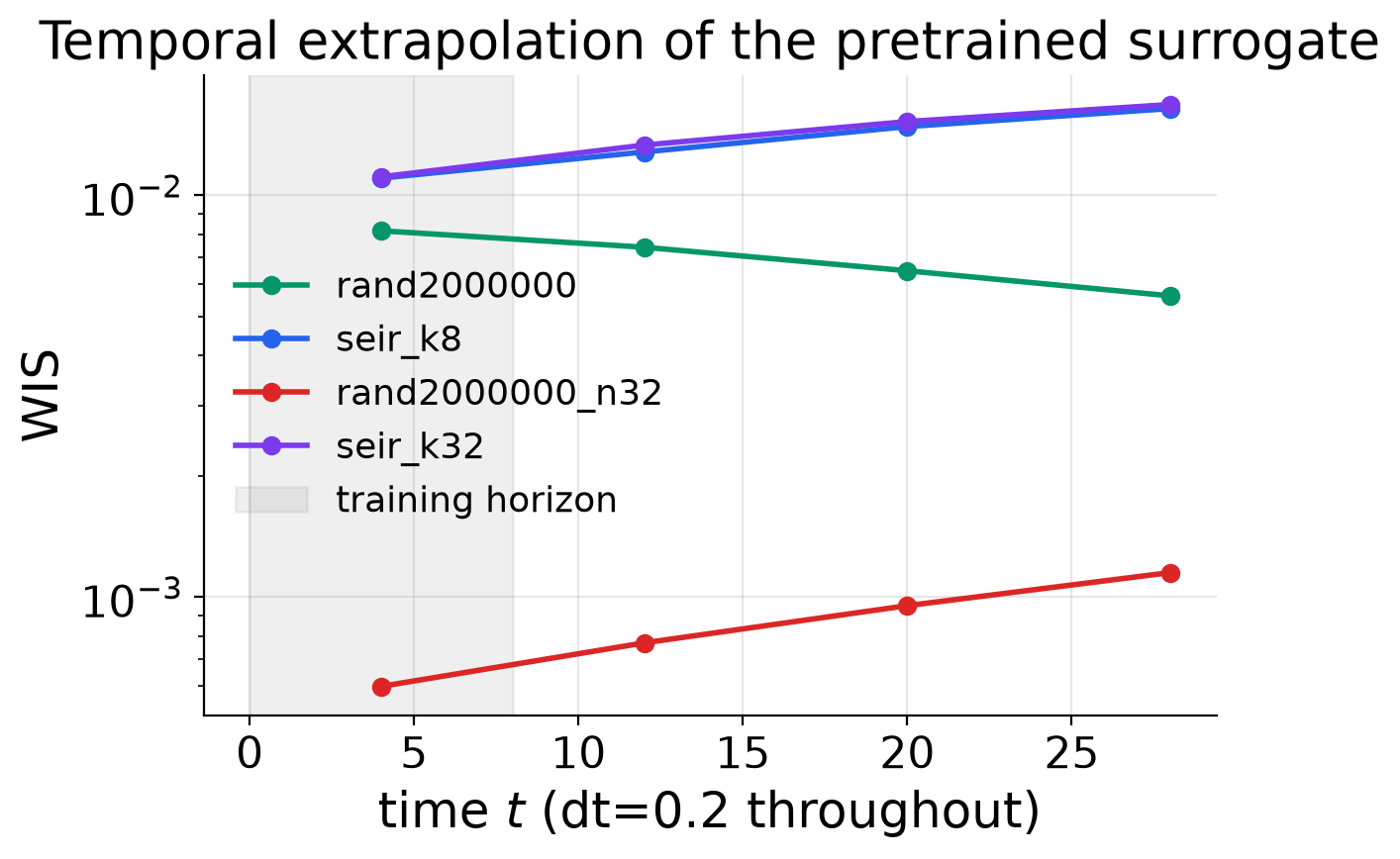}
    \caption{Horizon extrapolation ($4\times$).}
  \end{subfigure}
  \begin{subfigure}{0.4\textwidth}\centering
    \includegraphics[width=\linewidth]{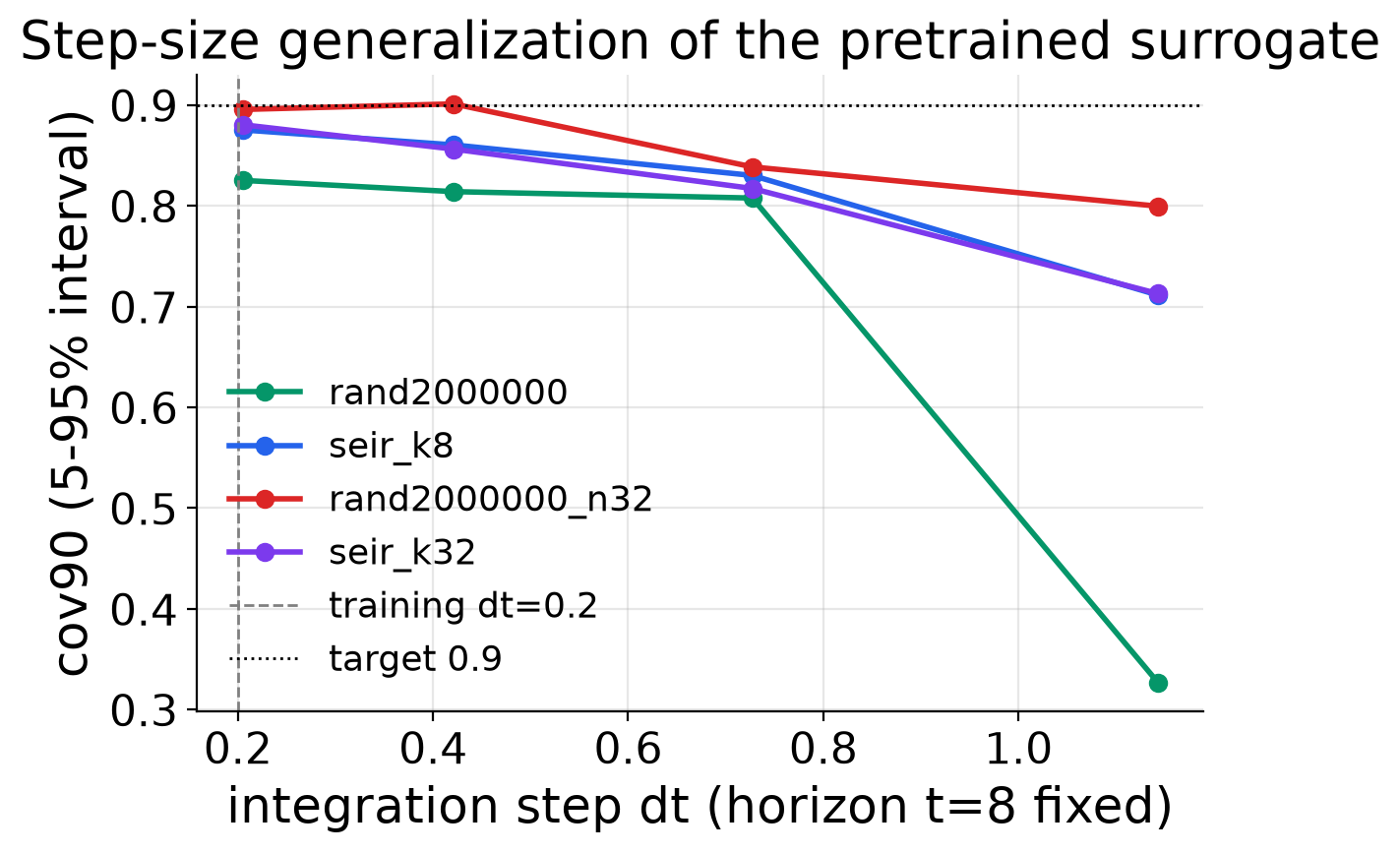}
    \caption{Step-size coarsening (cov90).}
  \end{subfigure}
  \caption{Temporal extrapolation and step-size coarsening of the frozen surrogate.}
  \label{fig:temporal}
\end{figure}

The learned field is time-independent, so the pretrained surrogate can be integrated
beyond its trajectories times seen during training with no retraining. We consider two extensions in time. \emph{Horizon:} integrating to $t{=}32$ ($4\times$
the $t\le8$ training horizon) at the same $\mathrm{d}t$, WIS rises only $\sim1.3\times$ over the
window $(0,8]\to(24,32]$ and coverage drifts $0.87\to0.80$ (Fig.~\ref{fig:temporal}, left).
\emph{Step size:} evaluating at integration steps coarser than the $\mathrm{d}t{=}0.2$ training
grid, the median prediction is step-size invariant to $\sim4\times$ and calibration holds to
$\sim4\times$ before degrades at an impractically large step size (Fig.~\ref{fig:temporal}, right).

\subsection{Inverse: Parameter Recovery}
\begin{figure}[!ht]\centering
  \includegraphics[width=0.85\columnwidth]{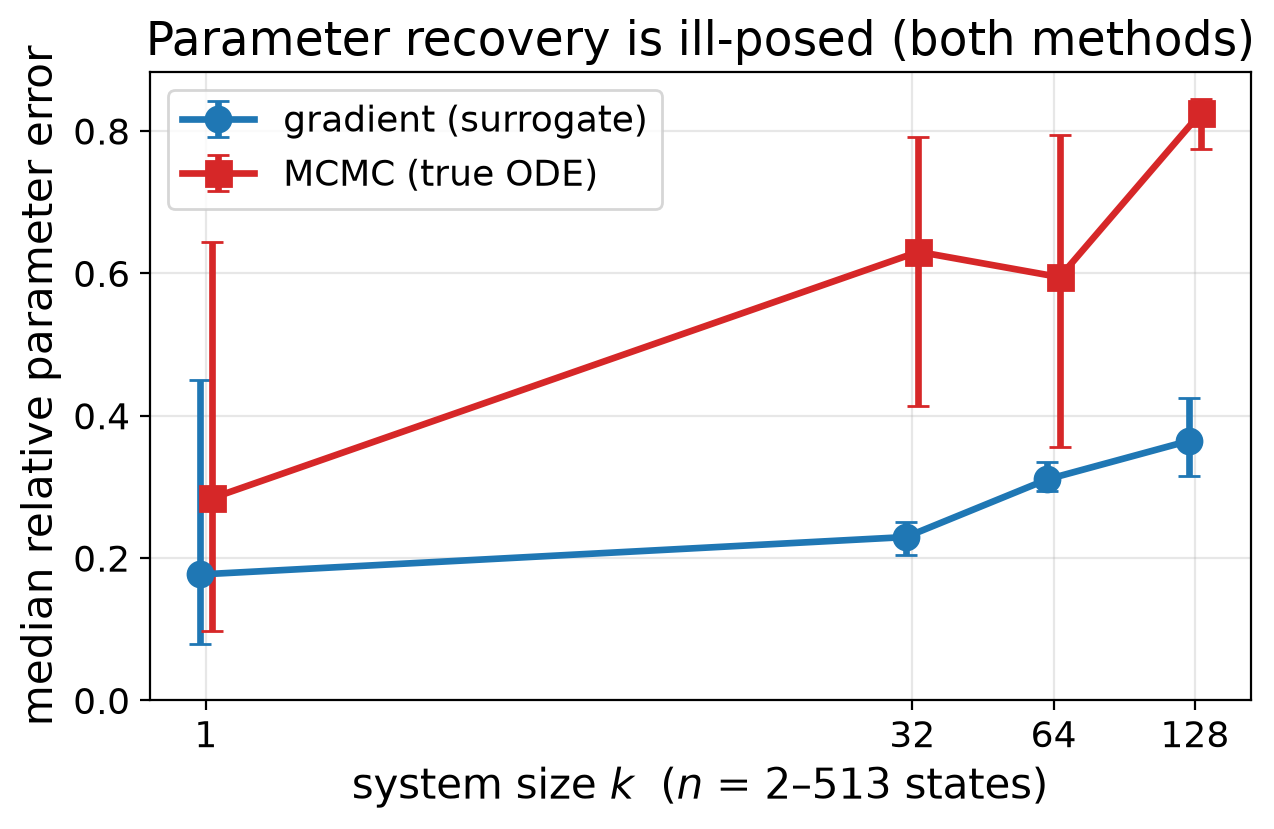}
  \caption{Median relative parameter-recovery error vs.\ system size (median $\pm$ IQR).}
  \label{fig:paramrec}
\end{figure}

The paper scores the \emph{predictive band}, not parameter recovery, as the systems are
unidentifiable. Figure~\ref{fig:paramrec} makes this explicit: the median relative
parameter error is large for \emph{both} the gradient inverse ($17$--$36\%$) and true-ODE MCMC
($28$--$83\%$), and grows with system size. Yet, the gradient inverse
recovers parameters better than MCMC at every size, and MCMC with limited budget degrades sharply at large $k$ -- the same random-walk mixing failure that collapses its predictive band. A parameter-recovery
comparison therefore scores an ill-posed quantity; the calibrated posterior-predictive band is
the well-defined target.

\subsection{Inverse: Compute Budgets and Baseline Details}
\begin{itemize}
  \item Gradient inverse: 300 Adam steps through the frozen surrogate (lr $0.05$), per instance.
  \item MCMC: random-walk Metropolis on the true ODE, Gaussian likelihood, flat prior on $\log\theta$; budget = burn + samples$\times$thin true-ODE solves = 1600 (small/test) vs 700 (large $k$) -- large-$k$ used \emph{fewer} iters, so its runtime is a lower bound. The reported MCMC runtime is the sequential single-chain sampling time (one true-ODE solve per step; the chain cannot be parallelized).
  \item Per-structure simulation-based inference (NPE/NRE): a normalizing-flow / ratio estimator with an MLP trajectory embedding, retrained per structure; NPE inference is slow at scale ($\sim1$\,h at $k{=}32$, $P{=}224$), NRE MCMC-based. Our single-pass encoder is a single general model (no per-structure training).
  \item FIM-ODE (foundation model, $13$M params): structure-blind, $D\le3$ only, point estimates; overtaken by the single-pass encoder as noise grows (Table~\ref{tab:fim}).
\end{itemize}

%% file: tab_batchtime.tex
\begin{table}[!ht]\centering\small
\begin{tabular}{llrrrrrr}
\toprule
method & $k$ & $B{=}1$ & $B{=}8$ & $B{=}32$ & $B{=}64$  & speedup \\
\midrule
single pass & 4 & 1.0\,ms & 0.7\,ms & 0.7\,ms & 0.7\,ms & $1.4\times$ \\
single pass & 32 & 1.4\,ms & 1.1\,ms & 1.1\,ms & 1.1\,ms & $1.3\times$ \\
\midrule
gradient & 4 & 58.6\,s & 8.2\,s & 2.5\,s & 1.5\,s &  $38\times$ \\
gradient & 32 & 60.7\,s & 8.5\,s & 4.6\,s & 4.4\,s &  $14\times$ \\
\bottomrule
\end{tabular}
\caption{Per-problem inference time vs.\ batch size $B$ (V100 GPU), for the two surrogate
inverses at a small ($k{=}4$) and large ($k{=}32$) system.}
\label{tab:batchtime}
\end{table}

%% file: tab_nonpoly.tex
\begin{table*}[!ht]\centering\small
\setlength{\tabcolsep}{6pt}
\begin{tabular}{l cc cc cc l}
\toprule
 & \multicolumn{2}{c}{swap (ours)} & \multicolumn{2}{c}{no-prod} & \multicolumn{2}{c}{no-prod\,+\,poly} & \\
\cmidrule(lr){2-3}\cmidrule(lr){4-5}\cmidrule(lr){6-7}
incidence $g(S,I)$ & WIS & cov90 & WIS & cov90 & WIS & cov90 & poly.\ pipeline \\
\midrule
sublinear $SI^{0.5}$            & $\mathbf{0.011}$ & $\mathbf{0.88}$ & $0.031$ & $0.36$ & $0.020$ & $0.63$ & norm+rescale ($\lambda{=}91$) \\
superlinear $SI^{1.5}$          & $\mathbf{0.004}$ & $\mathbf{0.81}$ & $0.031$ & $0.15$ & $0.077$ & $0.73$ & norm+rescale ($\lambda{=}811$) \\
saturating $SI/(1{+}8I)$        & $\mathbf{0.008}$ & $\mathbf{0.74}$ & $0.014$ & $0.63$ & $0.011$ & $0.65$ & recast only \\
Ricker $SI\,e^{-3I}$           & $\mathbf{0.010}$ & $\mathbf{0.73}$ & $0.012$ & $0.73$ & $0.013$ & $0.71$ & recast only \\
Hill $SI^{3}/(0.1^{3}{+}I^{3})$ & $\mathbf{0.014}$ & $\mathbf{0.83}$ & $0.035$ & $0.35$ & $144.5$ & $0.08$ & norm+rescale ($\lambda{=}3594$) \\
\bottomrule
\end{tabular}
\caption{\textbf{Generalized product vs.\ polynomialization.} Replacing SEIR incidence with-non polynomials. \emph{swap}
feeds the true nonlinearity $g(S,I)$ into the pretrained operator with no retraining. \emph{no-prod}
is the mass-action baseline: it has no product feature to swap. The
alternative is to \emph{polynomialize} $g$ into auxiliary states and run the no-product model
(\emph{no-prod\,+\,poly}), with the recast recipe in the last column. While polynomialization often helps, \emph{swap} achieves
better results.}
\label{tab:nonpoly}
\end{table*}

%% file: tab_jensen.tex
\begin{table}[t]\centering\small
\setlength{\tabcolsep}{8pt}
\begin{tabular}{cccc}
\toprule
$k$ & $\mathrm{sig}{=}0.4$ & $\mathrm{sig}{=}0.6$ & $\mathrm{sig}{=}0.8$ \\
\midrule
4  & $72\%$ & $68\%$ & $88\%$ \\
8  & $80\%$ & $78\%$ & $\mathbf{98\%}$ \\
16 & $88\%$ & $92\%$ & $\mathbf{97\%}$ \\
\bottomrule
\end{tabular}
\caption{\textbf{Beyond the median.} Fraction of held-out $\mathrm{SEIR}$ systems where the
surrogate's median trajectory is closer to the true median-of-solves than a single
$\mathrm{solve}(\mathrm{median}\,\theta)$, vs.\ system size $k$ and transmission-rate spread
$\mathrm{sig}$ ($n{=}60$/cell.}
\label{tab:jensen}
\end{table}